\documentclass[english]{article}
\usepackage[T1]{fontenc}
\usepackage[latin9]{inputenc}
\usepackage{geometry}
\usepackage{float}
\usepackage{multirow}
\usepackage{amsmath}
\usepackage{amsthm}
\usepackage{graphicx}
\usepackage{lipsum}

\makeatletter
\newcommand\blfootnote[1]{%
  \begingroup
  \renewcommand{\@makefntext}[1]{\noindent\makebox[1.8em][r]#1}
  \renewcommand\thefootnote{}\footnote{#1}%
  \addtocounter{footnote}{-1}%
  \endgroup
}
\makeatother

\makeatletter

\providecommand{\tabularnewline}{\\}
\floatstyle{ruled}
\newfloat{algorithm}{tbp}{loa}
\providecommand{\algorithmname}{Algorithm}
\floatname{algorithm}{\protect\algorithmname}

\theoremstyle{plain}
\newtheorem{thm}{\protect\theoremname}
\theoremstyle{plain}
\newtheorem{lem}[thm]{\protect\lemmaname}
\theoremstyle{remark}
\newtheorem{rem}[thm]{\protect\remarkname}

\@ifundefined{date}{}{\date{}}
\usepackage{caption}
\usepackage{array}
\usepackage{setspace}

\usepackage{algorithm}
\usepackage{algpseudocode}\usepackage{multirow}
\usepackage[affil-it]{authblk}
\usepackage{url}
\usepackage[colorlinks,allcolors=blue]{hyperref}
\usepackage{footmisc}
\usepackage[style=apa, natbib=true, hyperref=true]{biblatex}
\usepackage{xcolor}
\usepackage{amsfonts}

\title{A Novel Markov Model for Near-Term Railway Delay Prediction}

\author[1]{Jin Xu}
\author[2]{Weiqi Wang}
\author[3]{Zheming Gao \thanks{Corresponding author: gaozheming@ise.neu.edu.cn}}
\author[4]{Haochen Luo}
\author[4]{Qian Wu}

\affil[1]{School of Science and Engineering, the Chinese University of Hong Kong, Shenzhen, Guangdong 518172, China}

\affil[2]{Department of Mathematics and Statistics, Concordia University, Montreal, QC, H3G 1M8, Canada}

\affil[3]{College of Information Science and Engineering, Northeastern University, Shenyang, Liaoning 110819, China}

\affil[4]{Department of Industrial \& Systems Engineering, Texas A\&M University, College Station, TX 77843, USA}

\usepackage{babel}
\providecommand{\lemmaname}{Lemma}
\providecommand{\remarkname}{Remark}
\providecommand{\theoremname}{Theorem}

\@ifundefined{showcaptionsetup}{}{%
 \PassOptionsToPackage{caption=false}{subfig}}
\usepackage{subfig}
\makeatother

\usepackage{babel}
\providecommand{\lemmaname}{Lemma}
\providecommand{\remarkname}{Remark}
\providecommand{\theoremname}{Theorem}

\begin{document}

\maketitle 
\begin{abstract}
\blfootnote{\dag The first two authors contributed to the work equally and should be regarded as co-first authors.}
Predicting the near-future delay with accuracy for trains is momentous
for railway operations and passengers' traveling experience. This
work aims to design prediction models for train delays based on Netherlands
Railway data. We first develop a chi-square test to show that the
delay evolution over stations follows a first-order Markov chain.
We then propose a delay prediction model based on non-homogeneous
Markov chains. To deal with the sparsity of the transition matrices of the Markov chains,
we propose a novel matrix recovery approach that relies on Gaussian
kernel density estimation. Our numerical tests show that this recovery
approach outperforms other heuristic approaches in prediction accuracy. The Markov chain model we propose also shows to be better than
other widely-used time series models with respect to both interpretability
and prediction accuracy. Moreover, our proposed model does not require
a complicated training process, which is capable of handling large-scale
forecasting problems. 
\end{abstract}

\section{Introduction}

As one of the primary means of transport, railways provide freight
shipments and passenger services that enable a huge number of goods
and people to travel. Every year, about 10,000 billion freight tonne-kilometers
and 3,000 billion passenger-kilometers are traveled via railways around
the world (\cite{UICrailwaystat2015}). Punctuation is one of the most
crucial measures to quantify the quality of railway operations and
passengers' traveling experience. A train is punctual if it arrives
or departs at the planned time specified by the timetable. In the
ideal scenario, the trains are operated punctually as the timetable.
However, the railway operations would inevitably encounter disturbances,
and the punctuation of the railway system could be determined by various
factors, including the severe weather conditions, unexpected mechanical
failure, drivers' and travelers' behavior, and temporary speed restrictions
(\cite{olsson2004influencing,li2021near,nabian2019predicting}). As
shown in (\cite{harris2013improving}), the delay variations in Norway
by different railway lines are significant, with the best-performing
route achieving 94.4\% and the worst routes achieving only near 80\%,
against the target of 90\%. An accurate real-time delay predicting
model thus would help the railway operators better coordinate and
reschedule the trains, thereby improving the railway system reliability
and reducing system operating costs. Moreover, announcing accurate
delay information to passengers can better assist them in making travel
plans, which improves customers' traveling experience.

In this paper, we aim to develop a data-driven model that can predict
the delay of a train in the near future. To accurately predict the
delay is challenging for the following reasons. First, training a
large number of customized models could be challenging. For the same
train that travels through multiple stations, its delay at one station
may be distinct from another. For different trains operating at different
stations and regions, their delay processes could also be distinct.
The notable heterogeneity of delay's evolution across trains and stations
makes it impractical to train a single unified model that predicts
the delay for all the trains and stations. Therefore, building a customized
model for each train at a particular station is necessary. However,
training a unique model for each train at each station could be challenging
for a large area where many trains are operated. For instance, the
Netherlands Railways data (\cite{informsrail}) 
contains more than 6000 trains and 750000 stations. Even in the optimistic
case where each station has only one train to travel through, more
than 750000 models still need to be trained if we consider a unique
model for each station.

Second, there might be no adequate records of delays in the historical
data. For example, the worst route in (\cite{harris2013improving})
achieves an 80\% punctuation rate, which implies that most of the
historical data are still punctual. Moreover, the variety of delay
causes makes each type of delay even less recorded. It is 
challenging to train an accurate prediction model with such small historical
data of delays.

Third, some machine learning models are recently applied to delay
prediction, such as the artificial neural networks (ANN) (\cite{HUANG2020234}).
Although these models may do well in predicting the delay, they may
lack interpretability and cannot provide the insights of
delays to the railway operators. In addition, these models usually
require multiple data streams as input and many data to train. They
would be less accurate when certain streams of data are not recorded,
or when the training data are small. Moreover, these machine learning
models usually require much manual intervention in hyper-parameter
tuning. The long training and tuning processes make these models hard
to implement in scenarios where a large number of models need to
be built.

To overcome the challenges above, we propose a non-homogeneous Markov
chain model to predict the delay over stations. The main contributions
of our work are summarized as follows. 
\begin{itemize}
\item Markov chain modeling: We build our prediction model by assuming the
delay evolution over stations follows a non-homogeneous Markov chain.
Each transition matrix in our model can characterize the unique pattern
of delay evolution between two adjacent stations that a train travels
through. The delay evolution over multiple stations can be decently
captured by the Chapman-Kolmogorov equations of the Markov chain model. 
\item Transition matrix recovery method: We propose a Gaussian-kernel-based
method to recover the transition matrices for the Markov chain when
the training data are limited. This recovery method captures
the delay transition probabilities from the existing data. We show
that this recovery method achieves a higher prediction accuracy than
the other heuristic approaches. 
\item Markov property test: We propose a chi-square Markov property test
for the non-homogeneous Markov chain model when the transition matrix
is sparse due to a lack of training data. The Markov property test
verifies our assumption that the delay evolution over stations has
a first-order Markov property. It strengthens the interpretability
of our prediction model as well. 
\item Accuracy and lightweight: We conduct numerical experiments to verify
the effectiveness and efficiency of the proposed model in delay
prediction. The proposed model outperforms other commonly used time-series-based
prediction models with respect to accuracy. Besides, only a series
of transition matrices are necessary to be stored for each train in
prediction. Recovering these matrices needs less computational power
than other methods. Our model is thus suitable for forecasting the
delays in a railway system with a large number of trains and stations.
Moreover, our Markov chain model only relies on the delay data at
each station for training. So it can be applied in scenarios where
other factors related to train operation are not recorded. 
\end{itemize}
We organize the rest of the paper as follows. We review the related
work in Section \ref{sec:Related-Work} and describe the railway delay
forecasting problem in Section \ref{sec:Problem-Description}. In
Section \ref{sec:The-Markov-Chain}, we provide a detailed description
of the proposed Markov chain model. We test our model and compare
it with other benchmarks in Section \ref{sec:Forecasting-Results-and}.
We provide the conclusion and discuss our future research in Section
\ref{sec:Conclusion-and-Future}.


\section{Related Work}

\label{sec:Related-Work}

The models for traffic delay prediction have been investigated from
different perspectives. Some research focuses on the relationship
between railway delay and various factors in the railway systems.
For instance, \citet{olsson2004influencing} analyze
the correlation between train departure delay, number of passengers,
and occupancy ratio using Norwegian railway data. \citet{goverde2005punctuality}
adopts linear regression to explain the dependencies between train
services with a transfer connection and the impact of the bottleneck
in a particular station at Eindhoven. \citet{flier2009mining}
employ linear regression to analyze the delay dependencies on resource
conflicts and maintained connections using the Swiss Railways data. \citet{markovic2015analyzing} investigate support
vector regression (SVR) models that capture the relation between passenger
train arrival delays and various characteristics of a railway system,
such as the passenger train category, the scheduled time of arrival
at the station, the infrastructure influence, the percentage of the
journey completed distance-wise, and the traveling distance. All these
research studies mainly focus on the dependency between delay and
existing system characteristics. They aim to understand the factors
influencing railway delays so that to provide guidance to system operators
and planners. Performing real-time delay forecasting is not the main
focus of these studies.

Recently, machine learning approaches have been used for delay prediction. \citet{lessan2019hybrid} propose a Bayesian-network-based
train delay prediction model to characterize the complexity and dependency
nature of different train operations. \citet{yaghini2013railway}
use ANN models to predict monthly averaged passenger train delays
for Iranian railways. Real-time delay prediction is not the main focus
of this paper. \citet{oneto2018train} propose
shallow and deep extreme learning algorithms incorporating the types
of the running day (whether it is a weekday or holiday), dwell times,
and the running times for all the other trains running over the same
section of the railway network during the day, to predict the delays
in Italian railways. \citet{nabian2019predicting}
propose a random forest model by incorporating many features of the
railway system such as distance, number of stops, composite change,
and driver change. These models rely on features other than the
train's historical delay information for training, and may not be applicable in scenarios where these additional features are not recorded. Moreover, training these models could be time-consuming,
especially when the number of trains and stations is large.

Some statistical time series models that only rely on historical delay
information have been applied to traffic forecasting. \citet{suwardo2010arima} employ the Autoregressive Integrated
Moving Average (ARIMA) model to predict bus travel time on the Ipoh-Lumut
corridor in Malaysia. \citet{lippi2013short} compare
the statistical time series models with the learning-based models
on short-term traffic flow forecasting based on data collected from
the California Freeway Performance Measurement System. The seasonal
ARIMA (SARIMA) model with Kalman filter turns out to be the most accurate
one. Although the ARIMA models are applicable in many scenarios,
they may not perform well in railway delay forecasting where the delay evolution over stations is significantly heterogeneous and complex.

Recently, hybrid statistical time series and machine learning
models have been proposed for traffic forecasting. \citet{zhang2003time}
proposes a hybrid ARIMA and neural network model for time series forecasting. \citet{ma2020hybrid} concatenate a fundamental neural
network model with the ARIMA model for network-wide traffic forecasting. \citet{ge2021arima} propose a hybrid ARIMA model
and fuzzy SVR for high-speed railway passenger traffic forecasting.
However, these learning-based models may lack interpretability, and
some statistical assumptions made in these models are hard to justify
in practice.

As a model of a time-dependent system, the Markov
chain describes a sequence of random events where each event depends
only on the states attained in the previous several events. The order of a Markov chain specifies how many previous events that the current state depends on. Many real-world
processes can be illustrated by first-order Markov chain models where the current state only depends on the state in the previous event, such as
the wildlife migration \citep{Huang1977non-homogeneous}, the relative
risk of dementia \citep{yu2010nonstationary}, and the flight booking
process \citep{van2022inventory}. The analytical results and the corresponding
inferences of Markov chains are well studied in the literature \citep{dobrushin1956central1,dobrushin1956central2,aalen1978empirical,anily1987ergodicity}.
However, 
it is still unknown if the delay evolution over stations for railway
systems follows a Markov chain of a certain order.

In summary, there are very few railway delay forecasting models with
accuracy, lightweight, and interpretability altogether. Although the
Markov chain is used for prediction in multiple scenarios, it is still
unclear if the Markov chain model can capture the delay evolution of railway systems.
Therefore, in our study, we plan to investigate the Markov property
of delays in railway systems and design prediction models that rely
on Markov chains without training burdens.

\section{Problem Description\label{sec:Problem-Description}}

\subsubsection*{Delay Prediction}

Passenger train transportation is a very important mode of transport
in the Netherlands. Over a million passengers in Netherlands travel
by train every day. As the biggest passenger train operator, the Netherlands
Railways operates almost 6,000 trains daily. Over the recent years,
data regarding delays in the entire railway network have been collected
and made available to the operators. These data can be potentially
used to train delay prediction models useful in passenger information
systems and dispatching centers. An accurate prediction for the delays
of trains in the near future can be announced to passengers through
broadcasting or smartphone apps, which will be beneficial for passengers
in adjusting their travel plans, hence improving their traveling experience.
Moreover, an accurate estimate of the delay in the near future will enable
the railway operation to make timely dispatching adjustments and reschedules,
especially if the delay increases or decreases dramatically.

The objective of this research work is to develop a data-driven prediction
model that predicts the train's delay trend (decrease, equal, or increase),
the delay's jump property (delay increases or decreases for more
than two minutes), and the actual delay (in minutes) in the near future (e.g., 20 minutes). Moreover, we aim to develop a general
real-time prediction model with low computational costs, so that it is capable to predict the delay for a railway system with a large number of trains operated.

\subsubsection*{Data Description}

The data used for building our model is provided by ProRail, an
infrastructure manager responsible for track maintenance and coordinating
train operations \citep{informsrail}. The raw dataset contains railway
operation history from September 4, 2017 to December 9, 2017, and
it consists of a planned timetable and the realization data.

The planned timetable contains the planned arrival and departure times for each train at each station it travels through.
The timetable is the same for Mondays, Tuesdays, Thursdays, and Fridays.
On Wednesdays, extra trains operate on a busy part of the network
between Eindhoven and Amsterdam. The timetables for weekends are not
included due to altered operations on weekends.

The realization data contains the actual delay for each operated train
during the recorded period. It includes the realized departure and
arrival times (in seconds). Canceled trains are missing from the realization
data. We will mainly rely on the realization data to train our prediction
model.

Additional information, including the crew schedules, rolling stock
circulation, infrastructure data, and weather conditions, are also
provided in the dataset. However, our proposed
model does not rely on these additional features. Thus our model is robust and can be deployed in many other systems when these additional
variables are not recorded in history.

\section{The Markov Chain Prediction Model\label{sec:The-Markov-Chain}}

Before introducing our Markov chain prediction model, we first answer the questions of whether the delay evolution over stations
follows a Markov chain and whether the Markov chain has order one.
Answering these questions will help us understand how the delay at
the earlier stations influences that at later ones. Moreover, it will
provide theoretical support in explaining why the Markov chain we
propose works. Therefore, in this section, we first discuss the Markov
property of the railway delays in Section \ref{subsec:Markov-Property-for}.
We then introduce the proposed non-homogeneous Markov chain prediction model
in Section \ref{subsec:The-Non-homogeneous-Markov}. In Section \ref{subsec:The-Gaussian-Kernel-Method},
we introduce the Gaussian kernel matrix recovery method in the circumstances
where transition matrices for the Markov chain are sparse.

\subsection{Markov Property for Railway Delays\label{subsec:Markov-Property-for}}

In this subsection, we develop a chi-square test to verify that the
delay evolution over stations for each train follows a first-order
Markov chain. We first introduce the chi-square test for sparse transition
matrices and then use the Netherlands Railways data to show that the
Markov chain has order one.

\subsubsection{Markov Property Test}

We first introduce the concept of Markov property as follows. To facilitate
our discussion, we now focus on a particular train that travels through
multiple stations. We denote the delay at station $t$ by $D(t)$,
where $D(t)\in[-N,N]$ is a bounded integer random variable. We denote
the originated station as station $1$, and the destination station
as station $M$. Assume $d(1),d(2),\dots,d(t)$ is a delay series
(rounded in minutes) from station $1$ to station $t$ ($t\leq M$).
The delays evolution over stations satisfies a $k^{th}$ order Markov
property if it satisfies

\begin{equation}
\boldsymbol{P}\left(D(t)=d(t)|D(t-j)=d(t-j),\forall j=1,\dots,t-1\right)=\boldsymbol{P}(D(t)=d(t)|D(t-j)=d(t-j),\forall j=1,\dots,k),\label{eq: kth order Markov property}
\end{equation}
where $k=1,\dots,t-1$. In particular, the delays satisfy a zero order
Markov property if

\begin{equation}
\boldsymbol{P}(D(t)=d(t)|D(t-j)=d(t-j),\forall j=1,\dots,t-1)=\boldsymbol{P}(D(t)=d(t)),\label{eq: zero order Markov property}
\end{equation}
When the zero-order Markov property holds, $D(t)$ is a random variable
independent of the random variables $D(t-1),\dots,D(1)$, which means
that the current delay is independent of the historical delays.

We aim to use a chi-square test to verify that the delay evolution
over stations follows a first-order Markov chain. The chi-square test
will be performed on the historical data. Therefore, before describing
how the chi-square test works, we first introduce some necessary notations
to describe the values obtained from the historical data: 
\begin{itemize}
\item $n_{j}(t)$: the number of observations in the historical data that
the train delays for $j$ minutes at station $t$; 
\item $n_{i,j}(t)$: the number of observations that the train delays for
$j$ minutes at station $t$, and delays for $i$ minutes at station
$t-1$; 
\item $n_{h,i,j}(t)$: the number of observations that the train delays
for $j$ minutes at station $t$, delays for $i$ minutes at station
$t-1$, and delays for $h$ minutes at station $t-2$. 
\end{itemize}
Based on the number of observations defined above, we derive the following
relations, which would be useful for future analysis:

\begin{eqnarray}
 &  & \begin{cases}
n_{j}(t)=\sum_{i=-N}^{N}n_{i,j}(t)=\sum_{h=-N}^{N}\sum_{i=-N}^{N}n_{h,i,j}(t),\\
n_{i}(t-1)=\sum_{j=-N}^{N}n_{i,j}(t)=\sum_{h=-N}^{N}\sum_{j=-N}^{N}n_{h,i,j}(t),\\
n_{h}(t-2)=\sum_{i=-N}^{N}\sum_{j=-N}^{N}n_{h,i,j}(t).
\end{cases}\label{eq:3-1}
\end{eqnarray}
Using the defined terms in Equations (\ref{eq:3-1}), we now define
the maximum likelihood estimate of the state transition probabilities
that will be used in chi-square test as follows: 
\begin{itemize}
\item $\hat{p}_{j}(t)=\frac{n_{j}(t)}{\sum_{l}n_{l}(t)}$: the frequency
that the train delays for $j$ minutes at station $t$. The value
$\hat{p}_{j}(t)$ is the \emph{maximal likelihood estimate} (MLE)
of the probability $p_{j}(t)=\boldsymbol{P}(D(t)=j)$. 
\item $\hat{p}_{i,j}(t)=\frac{n_{i,j}(t)}{\sum_{l}n_{i,l}(t)}$: the frequency
that the train delays for $j$ minutes at station $t$, given that
the train delays for $i$ minutes at station $t-1$. The value $\hat{p}_{i,j}(t)$
is the MLE of the probability $p_{i,j}(t)=\boldsymbol{P}(D(t)=j|D(t-1)=i).$ 
\item $\hat{p}_{h,i,j}(t)=\frac{n_{h,i,j}(t)}{\sum_{l}n_{h,i,l}(t)}$: the
frequency that the train delays for $j$ minutes at station $t$,
given that the train delays for $i$ minutes at station $t-1$ and
$h$ minutes at station $t-2$. The value $\hat{p}_{h,i,j}(t)$ is
the MLE of the probability $p_{h,i,j}(t)=\boldsymbol{P}(D(t)=j|D(t-1)=i,D(t-2)=h).$ 
\end{itemize}
After introducing the necessary notations above, we now perform the
Markov property test following a similar procedure as \citep{tan2002markov,bickenbach2003evaluating}. The general idea of the testing procedure is that we test the Markov
property from order zero, until a certain order of Markov property
is accepted.
However, it is important to note that the tests introduced in \citep{tan2002markov,bickenbach2003evaluating}
are for homogeneous Markov chains, i.e., the values of $p_{j}(t)$,
$p_{i,j}(t)$, and $p_{h,i,j}(t)$ do not vary for different stations
$t$. We relax this assumption in our work by modeling the delay evolution
as non-homogeneous Markov chains since the delay may have distinct
forms of evolution over stations. There are multiple reasons for the
delay to be non-homogeneous over stations. For instance, due to the
delay, the train can be overtaken by a slower train at a certain track.
As a result, this train cannot go faster than the slower train that
is now in front, which causes the delay to accumulate \citep{lee2016delay}.
Moreover, the traveling distance between stations could be distinct.
Two stations with a long distance in between may also have a large
slack time in the timetable, and the slack time can be used to reduce
the delay. The third reason could be the change of the rolling stock
composition at certain stations \citep{informsrail}, which may result
in delay distributions different from those stations without rolling
stock change.

A challenge for testing the Markov property for non-homogeneous Markov
chains is that the frequency is an approximation of the actual probability.
If the probability is small, for example, if $p_{i,j}\ll1$, it is
quite likely to observe $n_{i,j}=0$ from historical data. Then the obtained transition matrix
of the Markov chain that indicates the probabilities of transiting
from historical delays to the current delay may be sparse. To overcome
this challenge, we propose a Markov property testing approach for
the non-homogeneous Markov chain based on the likelihood ratio and
chi-square statistic by removing the zero rows and columns of the transition matrices. The likelihood ratio and chi-square statistic
are tested against a chi-square distribution whose degree of freedom
relies only on the non-zero empirical probabilities calculated from
the historical data.

We now test the null hypothesis that the Markov
chain has zero order for a specific station $t$, i.e., $H_{0}^{(0)}:\{\forall i,j:p_{i,j}(t)=p_{j}(t)\}$.
We define $\mathcal{A}(t)=\{j:n_{j}(t)>0\}$ as the index set
of the delay minutes observed from the historical data. Similarly,
we define $\mathcal{B}_{i}(t)=\{j:n_{i,j}(t)>0\}$ and $\mathcal{C}_{j}(t)=\{i:n_{i,j}(t)>0\}$.
Considering all $\hat{p}_{i,j}(t)$ as parameters testing the null
hypothesis, we then obtain the zero-order likelihood ratio \citep{koch1988parameter}
$LR^{(0)}(t)$ and chi-square statistic \citep{pearson1900x} $Q^{(0)}(t)$
for hypothesis tests as follows:

\begin{eqnarray}
LR^{(0)}(t) & = & -2\ln{\frac{\prod_{j}(\hat{p}_{j}(t))^{n_{j}(t)}}{\prod_{i,j}(\hat{p}_{i,j}(t))^{n_{i,j}(t)}}}=2\sum_{i,j:\hat{p}_{i,j}(t)\neq0}n_{i,j}(t)\ln\frac{\hat{p}_{i,j}(t)}{\hat{p}_{j}(t)},\label{eq:1}\\
Q^{(0)}(t) & = & \sum_{i,j:\hat{p}_{i,j}(t)\neq0}n_{i}(t-1)\frac{(\hat{p}_{i,j}(t)-\hat{p}_{j}(t))^{2}}{\hat{p}_{j}(t)}.\label{eq:2}
\end{eqnarray}

Then we analyze the degree of freedom for likelihood ratio and chi-square
tests in the following lemma. 
\begin{lem}
\label{lem:Both--and}Both $LR^{(0)}(t)$ and $Q^{(0)}(t)$ follow
an asymptotic chi-square distribution with degree of freedom 
\begin{eqnarray}
\left(|\mathcal{A}(t-1)|-1\right)\left(|\mathcal{A}(t)|-1\right).\label{eq:3}
\end{eqnarray}
\end{lem}

\begin{proof}
It has been proven in \citet{anderson1957statistical,bickenbach2003evaluating}
that $LR^{(0)}(t)$ and $Q^{(0)}(t)$ follow the asymptotic chi-square
distributions with identical degree of freedom. We thus consider the
degree of freedom for the chi-square distribution focusing on $Q^{(0)}(t)$.

Using a similar argument in \citet{anderson1957statistical}, we can
show that $\sum_{j:\hat{p}_{i,j}(t)\neq0}n_{i}(t-1)\frac{(\hat{p}_{i,j}(t)-\hat{p}_{j}(t))^{2}}{\hat{p}_{j}(t)}$
has an asymptotic chi-square distribution with a maximal degree of
freedom $|\mathcal{A}(t)|-1$. To further derive the degree of freedom
of $Q^{(0)}(t)$, we only need to compute how many rows of $i$ such
that $n_{i}(t)\neq0$. Here, we define the indicator function as 
\[
\boldsymbol{1}_{A}=\begin{cases}
1 & \text{if condition A is true,}\\
0 & \text{if condition A is false}.
\end{cases}
\]

Since $|\mathcal{A}(t)|=\sum_{j=-N}^{N}\boldsymbol{1}_{n_{j}(t)\neq0}$,
based on Equations (\ref{eq:3-1}), we have $|\mathcal{A}(t)|\geq\sum_{j=-N}^{N}\boldsymbol{1}_{n_{i,j}(t)\neq0}$
for any $i\in[-N,N]$. Now we compute the number of non-zero $\mathcal{B}_{i}$s
as follows: 
\begin{eqnarray*}
\sum_{i=-N}^{N}\boldsymbol{1}_{|\mathcal{B}_{i}(t)|\neq0} & = & \sum_{i=-N}^{N}\boldsymbol{1}_{\sum_{j=-N}^{N}n_{i,j}(t)\neq0}\\
 & = & \sum_{i=-N}^{N}\boldsymbol{1}_{n_{i}(t-1)\neq0}\\
 & \leq & |\mathcal{A}(t-1)|.
\end{eqnarray*}
Therefore, there is no more than $|\mathcal{A}(t-1)|$ number of $\mathcal{B}_{i}$.
Moreover, since $\sum_{j=-N}^{N}p_{j}(t)=1$, we should also subtract
$\left(|\mathcal{A}(t)|-1\right)$ number of degree of freedom from the summation.
Therefore, the degree of freedom under $H_{0}^{(0)}$ is given by
\[
\left|\mathcal{A}(t-1)\right|\left(|\mathcal{A}(t)|-1\right)-(|\mathcal{A}(t)|-1)=\left(|\mathcal{A}(t-1)|-1\right)\left(|\mathcal{A}(t)|-1\right).
\]
\end{proof}
\begin{rem}
In the scenarios where we do not have enough training data, it is
likely that $n_{i,j}(t)$ has rows and columns with all the elements
being zero. These zero rows and columns do not provide additional
information in the test, thus are removed when we calculate the degree
of freedom in Lemma \ref{lem:Both--and}. The degree of freedom given
in Lemma \ref{lem:Both--and} is thus the one we compute based on
the truncated matrix $n_{i,j}(t)$. It is the actual degree of freedom
for the matrix after removing all the zero rows and columns. Figure
\ref{fig:Truncated-Matrix} provides a demonstrative graph of the
truncated matrix.

\begin{figure}
\begin{centering}
\includegraphics[scale=0.7]{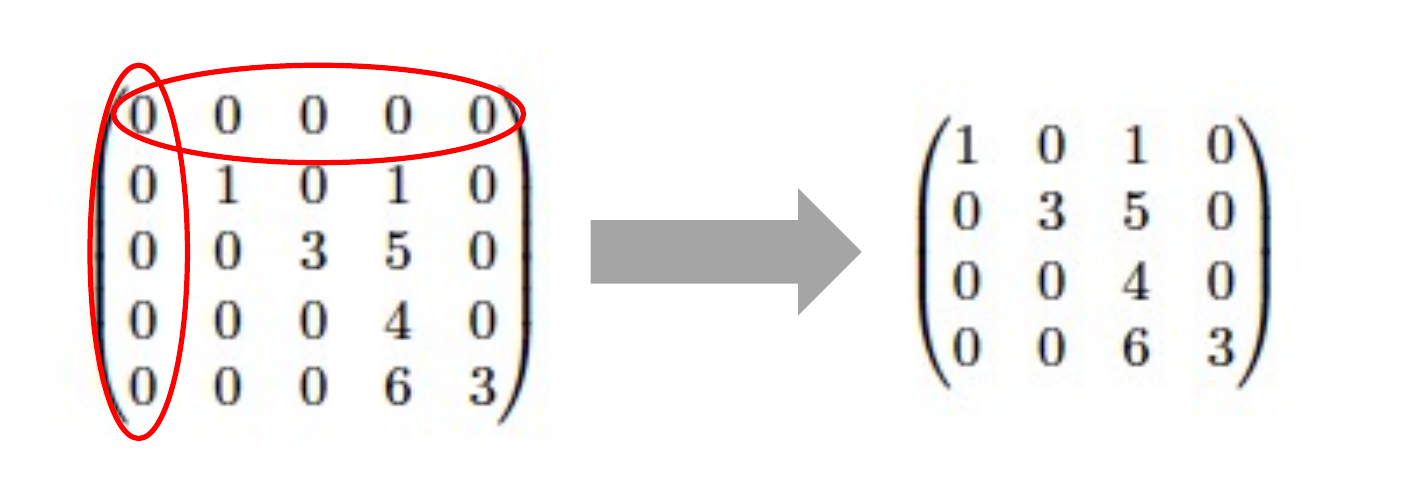}
\par\end{centering}
\caption{Matrix truncation. The left matrix has zero rows and columns, and
the right one is the truncated matrix after removing zero rows and
columns\label{fig:Truncated-Matrix}}
\end{figure}
\end{rem}

If the zero-order hypothesis is rejected, we further test the hypothesis
that the Markov chain has order one for a specific station $t$, i.e.,
$H_{0}^{(1)}:\{\forall h,i,j:p_{h,i,j}(t)=p_{i,j}(t)\}$. Similarly,
we obtain the first-order likelihood ratio $LR^{(1)}(t)$ and chi-square
statistic $Q^{(1)}(t)$ as follows:

\begin{eqnarray}
LR^{(1)}(t) & = & 2\sum_{h,i,j:\hat{p}_{h,i,j}(t)\neq0}n_{h,i,j}(t)\ln\frac{\hat{p}_{h,i,j}(t)}{\hat{p}_{i,j}(t)},\label{eq:4}\\
Q^{(1)}(t) & = & \sum_{h,i,j:\hat{p}_{h,i,j}(t)\neq0}n_{h,i}(t-1)\frac{(\hat{p}_{h,i,j}(t)-\hat{p}_{i,j}(t))^{2}}{\hat{p}_{i,j}(t)}.\label{eq:5}
\end{eqnarray}

Then we have the following lemma for the chi-square distribution that
these two statistics test against. 
\begin{lem}
\label{lem:We-have-that}Both $LR^{(1)}(t)$ and $Q^{(1)}(t)$ follow
the asymptotic chi-square distribution with degree of freedom 
\begin{equation}
\left(|\mathcal{A}(t-2)|-1\right)\left|\mathcal{A}(t-1)\right|\left(|\mathcal{A}(t)|-1\right).\label{eq:6}
\end{equation}
\end{lem}

\begin{proof}
We prove this lemma following a similar argument to the proof for
Lemma \ref{lem:Both--and} by focusing on the degree of freedom for
$Q^{(1)}(t)$. For each fixed $h$ and $i$, we have $\sum_{j=-N}^{N}\boldsymbol{1}_{n_{h,i,j}(t)\neq0}\leq|\mathcal{A}(t)|$,
the maximum degree of freedom is then $|\mathcal{A}(t)|-1$. We now
consider the number of $h$ and $i$ such that $n_{h,i}(t-1)\neq0$,
then $n_{h,i,j}(t)\leq n_{h,i}(t-1)$. Summing over these $h$ and
$i$, we have 
\begin{eqnarray*}
\sum_{h=-N}^{N}\sum_{i=-N}^{N}\boldsymbol{1}_{n_{h,i,j}(t)\neq0} & \leq & \sum_{h=-N}^{N}\sum_{i=-N}^{N}\boldsymbol{1}_{n_{h,i}(t-1)\neq0}\\
 & \leq & \sum_{h=-N}^{N}\boldsymbol{1}_{\sum_{i=-N}^{N}n_{h,i}(t-1)\neq0}|\mathcal{A}(t-1)|\\
 & \leq & \sum_{h=-N}^{N}\boldsymbol{1}_{n_{h}(t-2)\neq0}|\mathcal{A}(t-1)|\\
 & \leq & \left|\mathcal{A}(t-2)\right|\left|\mathcal{A}(t-1)\right|.
\end{eqnarray*}
Now the total degree of freedom is $\left|\mathcal{A}(t-2)\right|\left|\mathcal{A}(t-1)\right|\left(|\mathcal{A}(t)|-1\right)$.
We subtract it by the number of degree of freedom $\left|\mathcal{A}(t-1)\right|\left|\mathcal{A}(t)-1\right|$
we lose by imposing $\sum_{j}p_{i,j}(t-1)=1$. Therefore, the degree
of freedom under $H_{0}$ is given by 
\[
\left|\mathcal{A}(t-2)\right|\left|\mathcal{A}(t-1)\right|\left(|\mathcal{A}(t)|-1\right)-|\mathcal{A}(t-1)|\left(|\mathcal{A}(t)|-1\right)=\left(|\mathcal{A}(t-2)|-1\right)\left|\mathcal{A}(t-1)\right|\left(|\mathcal{A}(t)|-1\right).
\]

Another way to prove the lemma is to follow a similar argument in
\citet{anderson1957statistical}: We consider the statistic 
\[
\chi_{i}^{2}=\sum_{h,j:\hat{p}_{h,i,j}(t)\neq0}n_{h,i}(t-1)\frac{(\hat{p}_{h,i,j}(t)-\hat{p}_{i,j}(t))^{2}}{\hat{p}_{i,j}(t)}.
\]

The statistic $\chi_{i}^{2}$ has the degree of freedom $\left(|\mathcal{A}(t-2)|-1\right)\left(|\mathcal{A}(t)|-1\right)$
due to $\sum_{j}\hat{p}_{h,i,j}(t)=1$ and $\sum_{j}\hat{p}_{i,j}(t)=1$.
Therefore, $Q^{(1)}(t)=\sum_{i:\sum_{h,j}\hat{p}_{h,i,j}(t)\neq0}\chi_{i}^{2}$
has the degree of freedom $(|\mathcal{A}(t-2)|-1)|\mathcal{A}(t-1)|(|\mathcal{A}(t)|-1)$.
Hence proved. 
\end{proof}
If the first-order hypothesis $H_{0}^{(1)}$ is rejected, we can then
test the next null hypothesis that the Markov chain has order two.
For conciseness, we do not provide the details here since the railway
delay can be captured by a first-order Markov chain, as we will show
in the numerical study later. Algorithm \ref{alg:Markov-Property-Test}
summarizes the detailed procedure for the Markov property test, where
we use the chi-square statistics as the testing statistics.

\begin{algorithm}
\begin{algorithmic}[1] \State{Obtain the number of observations
$n_{j}(t)$, $n_{i,j}(t)$, and $n_{h,i,j}(t)$ from historical data}
\State{Obtain the frequencies $\hat{p}_{j}(t)=\frac{n_{j}(t)}{\sum_{l}n_{l}(t)}$,
$\hat{p}_{i,j}(t)=\frac{n_{i,j}(t)}{\sum_{l}n_{i,l}(t)}$, and $\hat{p}_{h,i,j}(t)=\frac{n_{h,i,j}(t)}{\sum_{l}n_{h,i,l}(t)}$.}
\State{Compute $Q^{(0)}(t)=\sum_{i,j:\hat{p}_{i,j}(t)\neq0}n_{i}(t-1)\frac{(\hat{p}_{i,j}(t)-\hat{p}_{j}(t))^{2}}{\hat{p}_{j}(t)}.$}
\State{Let $\alpha_{1}$ be the error probability for the zero order
hypothesis $H_{0}^{(0)}$} \If{$Q^{(0)}(t)<\chi_{\alpha_{1},df0}^{2}$,
where degree of freedom $df0=\left(|\mathcal{A}(t-1)|-1\right)\left(|\mathcal{A}(t)|-1\right)$}
\State{Do not reject $H_{0}^{(0)}$ } \Else \State{Reject $H_{0}^{(0)}$}
\State{Compute $Q^{(1)}(t)=\sum_{h,i,j:\hat{p}_{h,i,j}(t)\neq0}n_{h,i}(t-1)\frac{(\hat{p}_{h,i,j}(t)-\hat{p}_{i,j}(t))^{2}}{\hat{p}_{i,j}(t)}.$}
\State{Let $\alpha_{2}$ be the error probability for the first
order hypothesis $H_{0}^{(1)}$} \If {$Q^{(1)}(t)<\chi_{\alpha_{2},df1}^{2}$,
where degree of freedom $df1=\left(|\mathcal{A}(t-2)|-1\right)\left|\mathcal{A}(t-1)\right|\left(|\mathcal{A}(t)|-1\right)$.}
\State{Do not reject $H_{0}^{(1)}$} \Else \State{Reject $H_{0}^{(1)}$}
\EndIf \EndIf \end{algorithmic}

\caption{Markov Property Test \label{alg:Markov-Property-Test}}
\end{algorithm}

\subsubsection{Numerical Results}

We now conduct the numerical test to show that the delay evolution
follows a first order Markov chain, based on all the trains that are
scheduled to operate during 8:00-8:20 and 12:00-12:20 in the historical
data. In our numerical test, we let $N=15$ since only very few delays
($0.58\%$) in the historical data are out of the range of $[-N,N]$
with $N=15$.

Note that a train may have different activities in the same station in the historical data. For example, in the historical data, \emph{``V''}
denotes the departure, \emph{``D''} denotes passage without stop,
\emph{``A''} denotes arrival, \emph{``KV''} denotes the departure
at a short stop, and \emph{``KA''} denotes the arrival at a short
stop. A train will have \emph{``V''} and \emph{``A''} (or \emph{``KV''} and \emph{``KA''}) at the same station. Since a train may stop at a station longer than scheduled, train
delays can be different at the arrival and departure epochs at the same station.
Thus, in our analysis, we regard the delays at the arrival and departure
at the same station as at two different stations.

We calculate the total number of stations that all the trains travel
through during these two periods and the number of stations that reject
the zero-order and first-order Markov test. We provide the test results
for both likelihood ratio $LR$ and chi-square statistics $Q$ in
Table \ref{tab:Markov-Property-Test}. From Table \ref{tab:Markov-Property-Test}
we find that both statistics reject the $H_{0}^{(0)}$ with a high
frequency and reject $H_{0}^{(1)}$ with a low frequency. This result
shows that the delay evolution over stations follows a first-order
Markov chain.

\begin{table}
\begin{centering}
\begin{tabular}{|c|c|c|c|c|}
\hline 
Time period  & Total Stations  & Statistics  & Stations Reject $H_{0}^{(0)}$  & Stations Reject $H_{0}^{(1)}$\tabularnewline
\hline 
\hline 
\multirow{2}{*}{8:00-8:20} & \multirow{2}{*}{1632} & $LR^{(i)}$  & 1468  & 2\tabularnewline
\cline{3-5} \cline{4-5} \cline{5-5} 
 &  & $Q^{(i)}$  & 1571  & 3\tabularnewline
\hline 
\multirow{2}{*}{12:00-12:20} & \multirow{2}{*}{1861} & $LR^{(i)}$  & 1630  & 1\tabularnewline
\cline{3-5} \cline{4-5} \cline{5-5} 
 &  & $Q^{(i)}$  & 1784  & 2\tabularnewline
\hline 
\end{tabular}
\par\end{centering}
\caption{Markov Property Test\label{tab:Markov-Property-Test}}
\end{table}

\subsection{Prediction Model}

\label{subsec:Forecasting-Model}

We now introduce how to use the Markov chain framework to predict
delay distributions, and then discuss how to recover the transition
matrices of the Markov chain from historical data.

\subsubsection{The Non-homogeneous Markov Chain Framework\label{subsec:The-Non-homogeneous-Markov}}

Since we have shown the delay evolution over stations follows a first-order
Markov chain in Section \ref{subsec:Markov-Property-for}, it is reasonable
to utilize the first-order Markov chain model to predict the delay
in the near future. We aim to predict the delay for trains in a future
station, given the delay at the current station. Without loss of generality,
we suppose that the train is currently located at station $S$ with
delay $d(S)$ minutes. We assume that from the timetable, the train
is supposed to arrive at station $T$ in the future. We aim to predict
the delay $d(T)$ at station $T$.

We predict the delays $d(t)$ using a $1\times(2N+1)$ dimension probability
vector $\mathbf{v}(t)$, where the $i^{th}$ element of $\mathbf{v}(t)$
(denoted as $v_{i}(t)$) represents the probability that the train
is delayed for $i$ minutes at station $t$ in our prediction model.
We then have $\sum_{i=-N}^{N}v_{i}(t)=1$ since $v_{i}(t)=\boldsymbol{P}(D(t)=i)$.
The delay at the current station $d(0)$ is a given number, as we
already know the current delay. So we have $v_{d(S)}(S)=1$ and $v_{i}(S)=S$
for $i\neq d(S)$. The transition matrix from station $t-1$ to station
$t$ that incorporates all the transition probabilities is given by
\begin{eqnarray*}
P(t) & = & \begin{pmatrix}p_{-N,-N}(t) & p_{-N,-N+1}(t) & ... & p_{-N,N}(t)\\
p_{-N+1,-N}(t) & p_{-N+1,-N+1}(t) & ... & p_{-N+1,N}(t)\\
...\\
p_{N,-N}(t) & p_{N,-N+1}(t) & ... & p_{N,N}(t)
\end{pmatrix}.
\end{eqnarray*}

The delay distribution $\mathbf{v}(S+1)$ at station $S+1$ is then
given as $\mathbf{v}(S+1)=\mathbf{v}(S)P(S+1)$. By induction, we
have the Chapman-Kolmogorov equation
\begin{equation}
\mathbf{v}(T)=\mathbf{v}(S)\prod_{t=S+1}^{T}P(t).\label{eq:probability distribution v(T)}
\end{equation}
We then obtain the probability distribution $\mathbf{v}(T)$ for the
delay at station $T$.

\subsubsection{The Gaussian-Kernel Method for Transition Matrix Recovery\label{subsec:The-Gaussian-Kernel-Method}}

We now need to recover the transition matrix $P(t)$ from the historical
data. As described in Section \ref{subsec:Markov-Property-for}, the
transition probability is recovered by the empirical probability $\hat{p}_{i,j}(t)=\frac{n_{i,j}(t)}{\sum_{l=-N}^{N}n_{i,l}(t)}$.
However, when the historical data are limited, it is possible that
for some $i$, we have $\sum_{l=-N}^{N}n_{i,l}(t)=n_{i}(t-1)=0$.
This scenario means that there is no observation from the historical
data that the train is delayed for $i$ minutes at station $t-1$.
In that way, if the delay at the current station $S$ is $i$ minutes
and $n_{i}(S-1)=0$, then it is impossible to predict the delay at
station $S+1$. A heuristic approach to revolve this issue is to let
$\hat{p}_{i,i}(t)=1$ and $\hat{p}_{i,j}(t)=0$ for $j\neq i$. This
recovery approach is to assume that the delay does not change over
stations, if we do not observe the delay value from historical data.
This approach is robust in some cases. However, it does not utilize
the statistical information of the observed data. We will show in
Section \ref{sec:Forecasting-Results-and} that this approach can
have a poor performance in practice. In this subsection, we propose
a matrix recovery method that utilizes the existing observations.
The general idea of the proposed matrix recovery approach is to utilize
the Gaussian kernel density estimate to recover a two-dimensional
distribution matrix and then normalize this matrix into a transition
matrix.

To recover the two-dimensional matrix, we first consider the joint
distribution for $D(t-1)$ and $D(t)$. We suppose $f_{D(t-1),D(t)}(x,y):\mathbb{R}\times\mathbb{R}\rightarrow\mathbb{R}$
is a two-dimensional probability density function for $D(t)$ and
$D(t-1)$ that satisfies $\iint_{\mathbb{R}^{2}}f_{D(t-1),D(t)}(x,y)dxdy=1$.
To estimate $f_{D(t-1),D(t)}(x,y)$, we suppose $(\mathbf{x}_{1}(t),\mathbf{x}_{2}(t),...,\mathbf{x}_{m}(t))$
are $m$ observations for $(D(t-1),D(t))$ from the historical data
with mean $\bar{\mathbf{x}}(t)$. We then estimate the probability
density function using the Gaussian kernel density estimate provided
in \citep{silverman2018density} as follows: 
\begin{eqnarray*}
\hat{f}_{t,h}(\mathbf{x}) & = & \frac{1}{mh^{2}|\Sigma|^{\frac{1}{2}}}\sum_{i=1}^{m}\Phi\bigg(h^{-2}(\mathbf{x}-\mathbf{x}_{i}(t))^{T}\Sigma^{-1}(\mathbf{x}-\mathbf{x}_{i}(t))\bigg),
\end{eqnarray*}
where $\Phi(\boldsymbol{u})=\frac{1}{2\pi}e^{-\frac{\boldsymbol{u}}{2}}$,
$\Sigma=\frac{1}{m-1}\sum_{i=1}^{m}(\mathbf{x}_{i}(t)-\bar{\mathbf{x}}(t))(\mathbf{x}_{i}(t)-\bar{\mathbf{x}}(t))^{T}$
is the covariance matrix, and $h$ is a hyper-parameter called window
width that determines the fitting smoothness. Since we are fitting
a two-dimensional distribution, the optimal window width is chosen
as $h=m^{-1/6}$ \citep{silverman2018density}.

However, if $D(t-1)$ and $D(t)$ are in perfect correlation (the
covariance between two random variables equals 1), the covariance
matrix $\Sigma$ is singular. $D(t-1)$ and $D(t)$ are shown to be
perfectly correlated when delay change is zero or constant. This phenomenon
frequently occurs when the size of the historical data set is small.
To resolve this issue, we may add an i.i.d. disturbance to the original
data, e.g., we let $(\tilde{\mathbf{x}}_{1}(t),\tilde{\mathbf{x}}_{2}(t),...,\tilde{\mathbf{x}}_{m}(t))=(\mathbf{x}_{1}(t)+\boldsymbol{\varepsilon}_{1},\mathbf{x}_{2}(t)+\boldsymbol{\varepsilon}_{2},...,\mathbf{x}_{m}(t)+\boldsymbol{\varepsilon}_{m})$
to be the modified data, where each $\boldsymbol{\varepsilon}_{i}$
is a independent two-dimensional disturbance randomly chosen within
$[-\epsilon,\epsilon]\times[-\epsilon,\epsilon]$ for a small $\epsilon$.
We can thus obtain a non-singular covariance matrix using the modified
data.

We now recover the transition matrix using the fitted Gaussian kernel
density estimate. Specifically, we denote $\tilde{P}(t)$ as the recovered
transition matrix for station $t$, and each element $\tilde{p}_{i,j}(t)$
within $\tilde{P}(t)$ is given as follows: 
\begin{eqnarray*}
\tilde{p}_{i,j}(t) & = & \frac{\hat{f}_{t,h}((i,j))}{\sum_{k=-N}^{N}\hat{f}_{t,h}((i,k))}.
\end{eqnarray*}

We can then predict the delay at station $T$ by substituting $P(t)$
with $\tilde{P}(t)$ in Equation (\ref{eq:probability distribution v(T)}).
We summarize our prediction model in Algorithm \ref{alg:The-Forecasting-Model}.

\begin{algorithm}
\begin{algorithmic}[1] \State{\textbf{Initialization:} Given
historical data $(\mathbf{x}_{1}(t),\mathbf{x}_{2}(t),...,\mathbf{x}_{m}(t))$
with $t=S+1,...,T$ and delay $d(S)$ at station S} \State{Let
$\mathbf{v}(S)=(0,...,\overset{d(S)}{1},...,0)$}. \For{t from
$S+1$ to $T$} \Function{$\hat{f}_{t,h}$}{$\mathbf{x}$}

\State{Randomly generate $m$ two-dimensional small perturbations
$(\boldsymbol{\varepsilon}_{1},\boldsymbol{\varepsilon}_{2},...,\boldsymbol{\varepsilon}_{m})$
from uniform distribution on $[-\epsilon,\epsilon]\times[-\epsilon,\epsilon]$.}
\State{Modified data $(\tilde{\mathbf{x}}_{1}(t),\tilde{\mathbf{x}}_{2}(t),...,\tilde{\mathbf{x}}_{m}(t))=(\mathbf{x}_{1}(t)+\boldsymbol{\varepsilon}_{1},\mathbf{x}_{2}(t)+\boldsymbol{\varepsilon}_{2},...,\mathbf{x}_{m}(t)+\boldsymbol{\varepsilon}_{m})$.}
\State{Data average $\bar{\mathbf{x}}(t)=\frac{1}{m}\sum_{i=1}^{m}\tilde{\mathbf{x}}_{i}(t)$}
\State{Let $\Phi(\boldsymbol{u})=\frac{1}{2\pi}e^{-\boldsymbol{u}/2}$,
$\Sigma=\frac{1}{m-1}\sum_{i=1}^{m}(\tilde{\mathbf{x}}_{i}(t)-\bar{\mathbf{x}}(t))(\tilde{\mathbf{x}}_{i}(t)-\bar{\mathbf{x}}(t))^{T}$,
and $h=m^{-1/6}$.} \State{\Return $\hat{f}_{t,h}(\mathbf{x})=\frac{1}{mh^{2}|\Sigma|^{\frac{1}{2}}}\sum_{i=1}^{m}\Phi\bigg(h^{-2}(\mathbf{x}-\tilde{\mathbf{x}}_{i}(t))^{T}\Sigma^{-1}(\mathbf{x}-\tilde{\mathbf{x}}_{i}(t))\bigg).$}
\EndFunction

\State{Let $\tilde{P}(t)$ be the transition matrix with $\tilde{p}_{i,j}(t)=\frac{\hat{f}_{t,h}((i,j))}{\sum_{k=-N}^{N}\hat{f}_{t,h}((i,k))}$.}
\State{Obtain the delay distribution $\mathbf{v}(t)$ at station
$t$ by $\mathbf{v}(t)=\mathbf{v}(t-1)\tilde{P}(t)$} \EndFor \State{\Return
$\mathbf{v}(T)$} \end{algorithmic}

\caption{The Markov Chain Prediction Model with Gaussian Kernel\label{alg:The-Forecasting-Model}}
\end{algorithm}

\section{Prediction Results and Discussions\label{sec:Forecasting-Results-and}}

This section presents the numerical tests and discussions for the
proposed model. We first introduce the performance measures in Section
\ref{subsec:Performance-Measures}, and then discuss the metrics for
prediction in Section \ref{subsec:Delay-Forecasting-Metrics}. In
Section \ref{subsec:Comparison-of-Matrix}, we investigate different
methods for recovering the transition matrix. We then compare the
proposed Markov chain model with other time series models in Section
\ref{subsec:Comparison-with-Other}.

\subsection{Performance Measures\label{subsec:Performance-Measures}}

We evaluate the performance of the prediction models by considering
their capability to predict delays at the predicted station: 
\begin{enumerate}
\item the train's delay trend (decrease, equal, or increase, compared with
the current delay); 
\item whether there is a delay jump (i.e., the predicted delay increases
or decreases for more than two minutes, compared with the current
delay); 
\item the minutes of delay. 
\end{enumerate}
We evaluate our model on all the $M$ trains that operate during a
randomly selected time window. The model will forecast delay trends,
delay jumps, and minutes of delay for each train (we will discuss
in detail how these metrics are extracted from the delay distribution
in Section \ref{subsec:Delay-Forecasting-Metrics}). The model's forecasting
scores are calculated based on the prediction results of all the selected
trains, as we shall show in the following.

\subsubsection*{Delay Trend Prediction Score}

Before introducing the delay trend predictions score, we first define
the following terms: 
\begin{itemize}
\item \emph{True Positive} ($TP_{IN}$): The number of trains whose delay
is predicted to increase and the delay actually increases. 
\item \emph{True Negative} ($TN_{IN}$): The number of trains whose delay
is not predicted to increase, but the actual delay increases. 
\item \emph{False Positive} ($FP_{IN}$): The number of trains whose delay
is predicted to increase and the delay does not occur in reality. 
\item \emph{False Negative} ($FN_{IN}$): The number of trains whose delay
is not predicted to increase, but the actual delay increases. 
\item \emph{Total number of trains}: $M=TP_{IN}+TN_{IN}+FP_{IN}+FN_{IN}$ 
\end{itemize}
We then define the positive predictive value for the increasing trend
prediction ($PPV_{IN}$) as

\begin{eqnarray*}
PPV_{IN} & = & \frac{TP_{IN}}{TP_{IN}+FP_{IN}},
\end{eqnarray*}
and the true positive rate for the increasing trend prediction ($TPR_{IN}$)
as

\begin{eqnarray*}
TPR_{IN} & = & \frac{TP_{IN}}{TP_{IN}+FN_{IN}}.
\end{eqnarray*}
We next evaluate the model's prediction performance in increasing
trend by considering the F1 score defined as 
\begin{eqnarray*}
F_{IN} & = & 2\frac{PPV_{IN}\times TPR_{IN}}{PPV_{IN}+TPR_{IN}}.
\end{eqnarray*}
The F1 score is ranged from 0 and 1, and a high F1 score indicates
high classification performance \citep{tharwat2020classification}.

Similarly, we can define the model's F1 score for predicting delay
decreasing $F_{DE}$ and predicting delay remaining equal $F_{EQ}$.
We thus use the F1 score 
\begin{eqnarray*}
F_{TR} & = & \frac{1}{3}\left(F_{IN}+F_{DE}+F_{EQ}\right)
\end{eqnarray*}
as the metric to measure the model's performance in predicting delay
trends.

\subsubsection*{Delay Jump Prediction Score}

Like the F1 score defined for delay trend prediction, we use the F1
score to evaluate the prediction score for delay jump.

\subsubsection*{Minutes of Delay Prediction Score}

We evaluate the prediction accuracy for minutes of delay using the
root weighted mean square error (RWMSE). Denote $\hat{d}_{i}$ as
the predicted delay and $d_{i}$ is the actual (realized) delay for
train $i$. We let the weight for the absolute delays $|d_{i}|$ of
0 and 1 minute be 0.2, and weight for all the other delays be 0.8.
Therefore, the RWMSE has a greater weight on large delays, which penalizes
more if the large delays are not accurately predicted. We then define
the root weighted mean square error as 
\begin{eqnarray}
RWMSE & = & \sqrt{\sum_{i=1}^{M}\left(w_{1}\boldsymbol{1}_{|d_{i}|\leq1}+w_{2}\boldsymbol{1}_{|d_{i}|>1}\right)\left|\hat{d}_{i}-d_{i}\right|},\label{eq:8}
\end{eqnarray}
where $w_{1}=\frac{0.2}{\sum_{i=1}^{M}\boldsymbol{1}_{|d_{i}|\leq1}}$
and $w_{2}=\frac{0.8}{\sum_{i=1}^{M}\boldsymbol{1}_{|d_{i}|>1}}$.
It is easy to verify that 
\begin{eqnarray*}
\sum_{i=1}^{M}(w_{1}\boldsymbol{1}_{|d_{i}|\leq1}+w_{2}\boldsymbol{1}_{|d_{i}|>1}) & = & \sum_{i=1}^{M}\left(\frac{0.2}{\sum_{i=1}^{M}\boldsymbol{1}_{|d_{i}|\leq1}}\boldsymbol{1}_{|d_{i}|\leq1}+\frac{0.8}{\sum_{i=1}^{M}\boldsymbol{1}_{|d_{i}|>1}}\boldsymbol{1}_{|d_{i}|>1}\right)=1.
\end{eqnarray*}
So that Equation (\ref{eq:8}) is a valid RWMSE.

\subsubsection*{Total Prediction Score}

We use the total prediction score provided in \citet{informsrail}
to evaluate the model's general prediction performance. The total
prediction score is a linear combination of $F_{TR}$, $F_{JP}$,
and RWMSE, which is given as 
\begin{eqnarray}
Score & = & 10F_{JP}+5F_{TR}-RWMSE.\label{eq:9}
\end{eqnarray}

The total prediction score in Equation (\ref{eq:9}) values the delay
jump prediction more than the trend prediction and RWMSE. The reason
is that in reality, being unable to predict the drastic delay change
may cause severe damage to both the railway system schedulers and
passengers. Note that the performance of our proposed model is insensitive
to the weights provided in Equation (\ref{eq:9}). As we will show
later, our proposed model outperforms other models in each of the
$F_{TR}$, $F_{JP}$, and RWMSE scores. 

\subsubsection*{Testing Data}

We evaluate the model performance based on two data sets from the
historical data. \emph{Test Set 1 }contains 174 trains operated during
8:00-8:20, November 7, 2017. The mean of the actual delay at the predicted
station in \emph{Test Set 1} is 1.82183, and the variance is 8.30911.
\emph{Test Set 2} contains 222 trains operated during 12:00-12:20,
November 9, 2017, with the mean of the actual delay at the predicted
station in \emph{Test Set 2} being 0.35211 and the variance being
0.96505. The delay in \emph{Test Set 1} is more divergent than that
in \emph{Test Set 2}. All of our models are trained based on the historical
data from September 4, 2017 to December 9, 2017, excluding these the
data on these two testing dates.

\subsection{Delay Prediction Metrics\label{subsec:Delay-Forecasting-Metrics}}

Algorithm \ref{alg:The-Forecasting-Model} introduced in Section \ref{subsec:Forecasting-Model}
returns a prediction of distribution $\mathbf{v}(T)$ for the
delay value $D(T)$. We then compare the approaches to obtain the
predicted delay trend, delay jump, and minutes of delay from the distribution
$\mathbf{v}(T)$.

We first define the mean, mode, and median for $D(T)$ as follows: 
\begin{itemize}
\item Mean: $Mean(D(T))=\sum_{i=-N}^{N}v_{i}(T)\cdot i.$ 
\item Mode: $Mode(D(T))=\arg\max_{i}\{v_{i}(T)\}.$ 
\item Median: $Median(D(T))=\min\{i:\sum_{j=-N}^{i}v_{i}(T)\geq\frac{1}{2}\}.$ 
\end{itemize}
We also define the probability of delay increasing, decreasing, remaining
equal, and delay jump as follows: 
\begin{itemize}
\item Probability of delay increasing: $\boldsymbol{P}(D(T)>d(S)|D(S)=d(S))=\sum_{j=d(S)+1}^{N}v_{i}(T).$ 
\item Probability of delay decreasing: $\boldsymbol{P}(D(T)<d(S)|D(S)=d(S))=\sum_{j=-N}^{d(S)-1}v_{i}(T).$ 
\item Probability of delay remaining equal: $\boldsymbol{P}(D(T)=d(S)|D(S)=d(S))=v_{d(S)}(T).$ 
\end{itemize}
Moreover, we define the probability of delay jump as 
\begin{itemize}
\item $\boldsymbol{P}(|D(T)-d(S)|\geq2|D(S)=d(S))=1-\sum_{i=d(S)-1}^{d(S)+1}v_{i}(t).$ 
\end{itemize}
We now select the best one to predict the delay trend, delay jump,
and minutes of delay based on these defined metrics.

\subsubsection*{Delay Trend Prediction}

We now compare four approaches to predict the delay trend based on
the mean value, mode, median, and probabilities. When using the mean
value to predict the delay trend, we will do the following: 
\begin{itemize}
\item Return ``increase'' if $Mean(D(T))-d(S)\geq1$; Return ``decrease''
if $Mean(D(T))-d(S)\leq-1$; Return ``equal'' otherwise. 
\end{itemize}
The way of using the mode and median to predict the delay trend is
similar to that of using the mean. When using the increasing/decreasing/equal
probability to predict delay trend, we do the following: 
\begin{itemize}
\item Return ``increase'' if $\sum_{i=d(S)+1}^{N}v_{i}(T)>\max\{\sum_{i=-N}^{d(S)-1}v_{i}(T),v_{d(S)}(T)\}$;
Return ``decrease'' if $\sum_{i=-N}^{d(S)-1}v_{i}(T)>\max\{\sum_{i=d(S)+1}^{N}v_{i}(T),v_{d(S)}(T)\}$;
Return ``equal'' otherwise. 
\end{itemize}

\subsubsection*{Delay Jump Prediction}

Similar to the delay trend prediction, we compare four approaches
to predict whether there is a delay jump based on the mean, mode,
median, and probabilities. When using the mean to predict delay jump,
we will 
\begin{itemize}
\item Return ``yes'' if $|Mean(D(T))-d(S)|\geq2$; Return ``no'' otherwise. 
\end{itemize}
We will use the same criterion when using the mode and median to predict
delay jump. When using the jump probability to predict delay jump,
we will 
\begin{itemize}
\item Return ``yes'' when the probability of delay jump $1-\sum_{i=d(S)-1}^{d(S)+1}v_{i}(t)\geq0.5$;
Return ``no'' otherwise. 
\end{itemize}

\subsubsection*{Metrics Comparison}

For the minutes of delay prediction, we compare the performance of
using mean, mode, and median at the predictor.

We present the performance score of using each metric in Table \ref{tab:Metrics-Selection-and}.
For delay trend prediction, we find that using the mean, mode, median,
and probability in prediction results in similar scores of $F_{TR}$.
Using the mode in prediction results in the lowest score for \emph{Test
Set 1}, and the highest score for \emph{Test Set 2}. Using delay increasing/decreasing/equal
probability to predict delay trend has the highest score for \emph{Test
Set 1}, but its performance for \emph{Test Set 2} ranks only the third.
We thus choose the median to predict the delay trend, as its performance
at both testing data sets ranks second among the four metrics.

We choose the jump probability to predict the delay jump, as its $F_{JP}$
scores for both testing sets are much higher than the other metrics.
Using the probability to predict the delay jump is also more reasonable
than using the other metrics since both large delay increase and decrease
are regarded as delay jumps. An example is that when the conditional
delay distribution $p_{i,j}(t)$ is Gaussian with a large tail, both
large delay decrease and increase have a high probability, but the
mean/mode/median value returns a delay minute in the middle, indicating
no delay jump.

We choose the mean value to predict the minutes of delay since the
mean value achieves the lowest RWMSE for both testing sets. The reason
is that the mean value can better characterize the central tendency
of the delay distribution, and the data with extremely large or small
delays are rarely observed in history.

\begin{table}
\begin{centering}
\begin{tabular}{|c|c|c|c|c|c|}
\hline 
Performance Measure  & Test Data  & Mean  & Mode  & Median  & Probability\tabularnewline
\hline 
\hline 
\multirow{2}{*}{$F_{TR}$} & Test Set 1  & 0.56934  & 0.54502  & 0.57231  & \textbf{0.58312}\tabularnewline
\cline{2-6} \cline{3-6} \cline{4-6} \cline{5-6} \cline{6-6} 
 & Test Set 2  & 0.60353  & \textbf{0.73044}  & 0.69949  & 0.66006\tabularnewline
\hline 
\multirow{2}{*}{$F_{JP}$} & Test Set 1  & 0.48387  & 0.38596  & 0.45902  & \textbf{0.56716}\tabularnewline
\cline{2-6} \cline{3-6} \cline{4-6} \cline{5-6} \cline{6-6} 
 & Test Set 2  & 0.48276  & 0.59259  & 0.51852  & \textbf{0.62500}\tabularnewline
\hline 
\multirow{2}{*}{$RWMSE$} & Test Set 1  & \textbf{2.88631}  & 3.17746  & 2.96964  & N/A\tabularnewline
\cline{2-6} \cline{3-6} \cline{4-6} \cline{5-6} \cline{6-6} 
 & Test Set 2  & \textbf{2.58319}  & 2.81596  & 2.74579  & N/A\tabularnewline
\hline 
\end{tabular}
\par\end{centering}
\caption{Metrics Comparison \label{tab:Metrics-Selection-and}}
\end{table}

\subsection{Comparison of Matrix Recovery Methods\label{subsec:Comparison-of-Matrix}}

In Section \ref{subsec:The-Gaussian-Kernel-Method}, we have proposed
a Gaussian kernel method to recover the zero elements due to a lack
of training data in the transition matrix of the Markov chain. We
now compare the following matrix recovery methods with the Gaussian
kernel approach.

\subsubsection*{Diagonal Filling}

Under the diagonal filling approach, we recover the transition matrix
$\tilde{P}(t)$ in the following way. 
\begin{itemize}
\item If $\sum_{l=-N}^{N}n_{i,l}(t)\neq0$, let $\tilde{p}_{i,j}(t)=\frac{n_{i,j}(t)}{\sum_{l=-N}^{N}n_{i,l}(t)}$. 
\item If $\sum_{l=-N}^{N}n_{i,l}(t)=0$, let $\tilde{p}_{i,i}(t)=1$ and
$\tilde{p}_{i,j}(t)=0$ for $j\neq i$. 
\end{itemize}
The idea of the diagonal filling approach is that we use the frequency
from the historical data to recover the transition probability. If
there is no observation that the train is delayed for $i$ minutes
at station $t-1$, i.e., $\sum_{l=-N}^{N}n_{i,l}(t)=0$, we assume
that the delay at the station $t$ is identical to the delay at the
previous station $t-1$.

\subsubsection*{Uniform Filling}

Under the uniform filling approach, we recover the transition matrix
$\tilde{P}(t)$ in the following way. 
\begin{itemize}
\item If $\sum_{l=-N}^{N}n_{i,l}(t)\neq0$, we let $\tilde{p}_{i,j}(t)=\frac{n_{i,j}(t)}{\sum_{l=-N}^{N}n_{i,l}(t)}$. 
\item If $\sum_{l=-N}^{N}n_{i,l}(t)=0$, we let $\tilde{p}_{i,j}(t)=\frac{1}{2N+1}$
for $j\in\{-N,...,N\}$. 
\end{itemize}
The idea of the uniform recovery is similar to that of the diagonal
filling approach, as both of them will use the frequency to recover
the transition probability. The difference is that when delay for
$i$ minutes at station $t-1$ is not observed from historical data,
under the uniform filling approach, we assume that the delay at station
$t$ is uniformly distributed within $\{-N,...,N\}$.

\subsubsection*{A Gaussian Regression Recovery Approach}

One phenomenon that we observe from the train delay data is that for
each $i$, the conditional probability $p_{i,j}(t)$ is likely to
be concentrated around $p_{i,i}(t)$, i.e., $p_{i,j}(t)$ is greater
when $j$ is close to $i$. This is because the delay jump between
two stations is quite rare, and the delay minutes are likely to be
similar between two stations. So we can assume that the distribution
$p_{i,j}(t)$ follows a Gaussian distribution for each $i$, and develop
the Gaussian regression recovery approach as follows. 
\begin{enumerate}
\item If $\sum_{l=-N}^{N}n_{i,l}(t)\neq0$, we let $\tilde{p}_{i,j}(t)=\frac{n_{i,j}(t)}{\sum_{l=-N}^{N}n_{i,l}(t)}$. 
\item For $i$ such that $\sum_{l=-N}^{N}n_{i,l}(t)\neq0$, calculate the
mean $\hat{\mu}_{i}=\frac{\sum_{l=-N}^{N}n_{i,l}(t)\cdot l}{\sum_{l=-N}^{N}n_{i,l}(t)}$
and standard deviation $\hat{\sigma}_{i}=\frac{\sum_{l=-N}^{N}n_{i,l}(t)(l-\hat{\mu}_{i})^{2}}{\sum_{l=-N}^{N}n_{i,l}(t)-1}$
for the fitted Gaussian distribution. 
\item Perform a linear regression based on the fitted mean $\hat{\mu}_{i}$
and standard deviation $\hat{\sigma}_{i}$. Obtain two regressed linear
functions 
\begin{eqnarray*}
\mu_{i} & = & \alpha+\beta\cdot i,
\end{eqnarray*}
and 
\begin{eqnarray*}
\sigma_{i} & = & \check{\alpha}+\check{\beta}\cdot i,
\end{eqnarray*}
where $\mu_{i}$ and $\sigma_{i}$ are the mean and standard deviation
for the $i^{th}$ row of $\tilde{P}(t)$, and $\alpha$, $\beta$,
$\check{\alpha}$, and $\check{\beta}$ are fitted parameters. 
\item For $i$ such that $\sum_{l=-N}^{N}n_{i,l}(t)=0$, let 
\begin{eqnarray*}
\hat{\mu}_{i} & = & \alpha+\beta\cdot i,
\end{eqnarray*}
and 
\begin{eqnarray*}
\hat{\sigma}_{i} & = & \check{\alpha}+\check{\beta}\cdot i.
\end{eqnarray*}
Then let $\tilde{p}_{i,j}(t)=\frac{g_{\hat{\mu}_{i},\hat{\sigma}_{i}}(j)}{\sum_{l=-N}^{N}g_{\hat{\mu}_{i},\hat{\sigma}_{i}}(l)}$,
where $g_{\mu,\sigma}(x)$ is a one-dimension probability density
function of the Gaussian distribution with mean $\mu$ and standard
deviation $\sigma$. 
\end{enumerate}

\subsubsection*{Comparison and Discussion}

\begin{figure}
\begin{centering}
\subfloat[Original Matrix]{\includegraphics[scale=0.3]{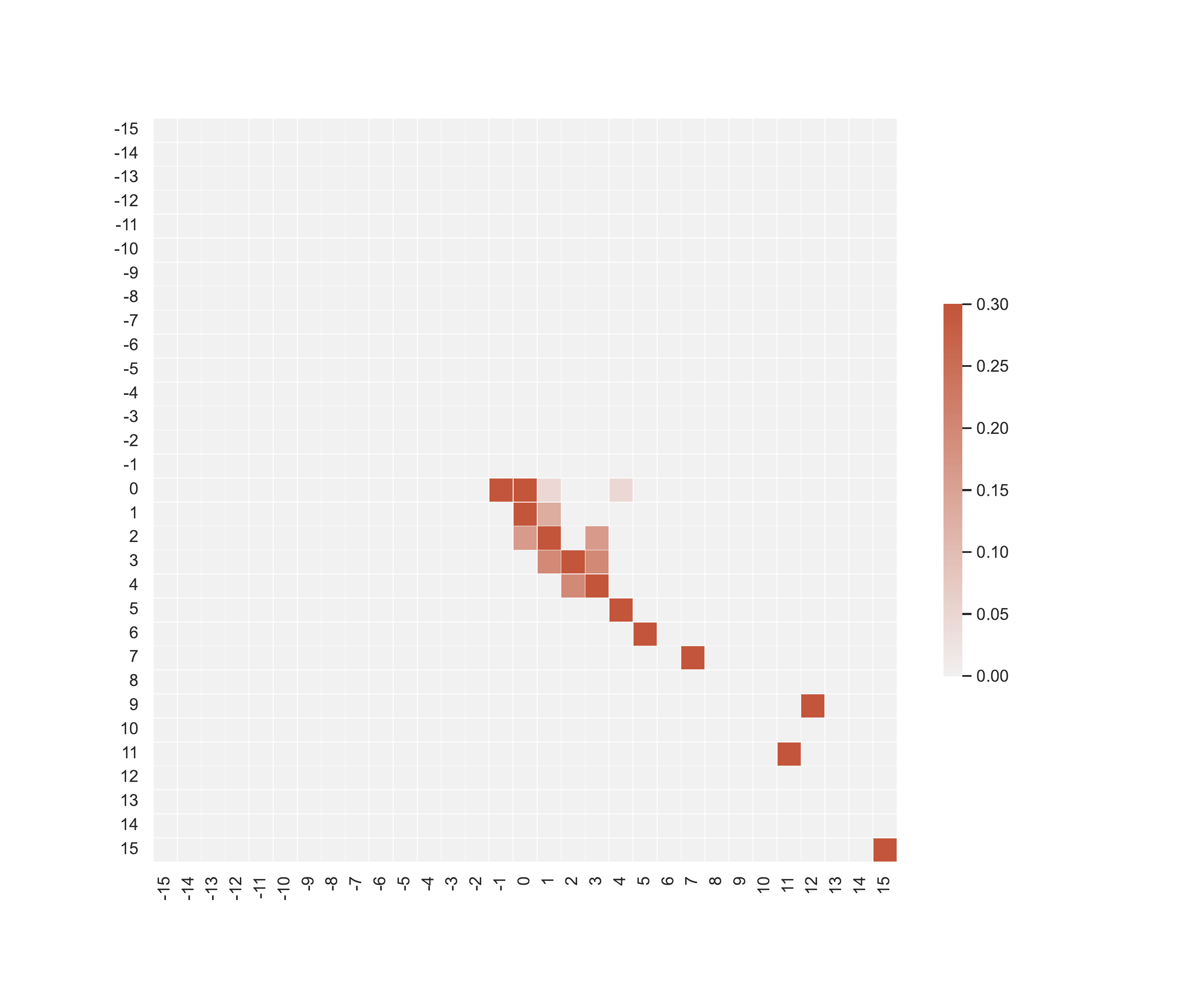}

}\subfloat[Joint Distribution]{\includegraphics[scale=0.3]{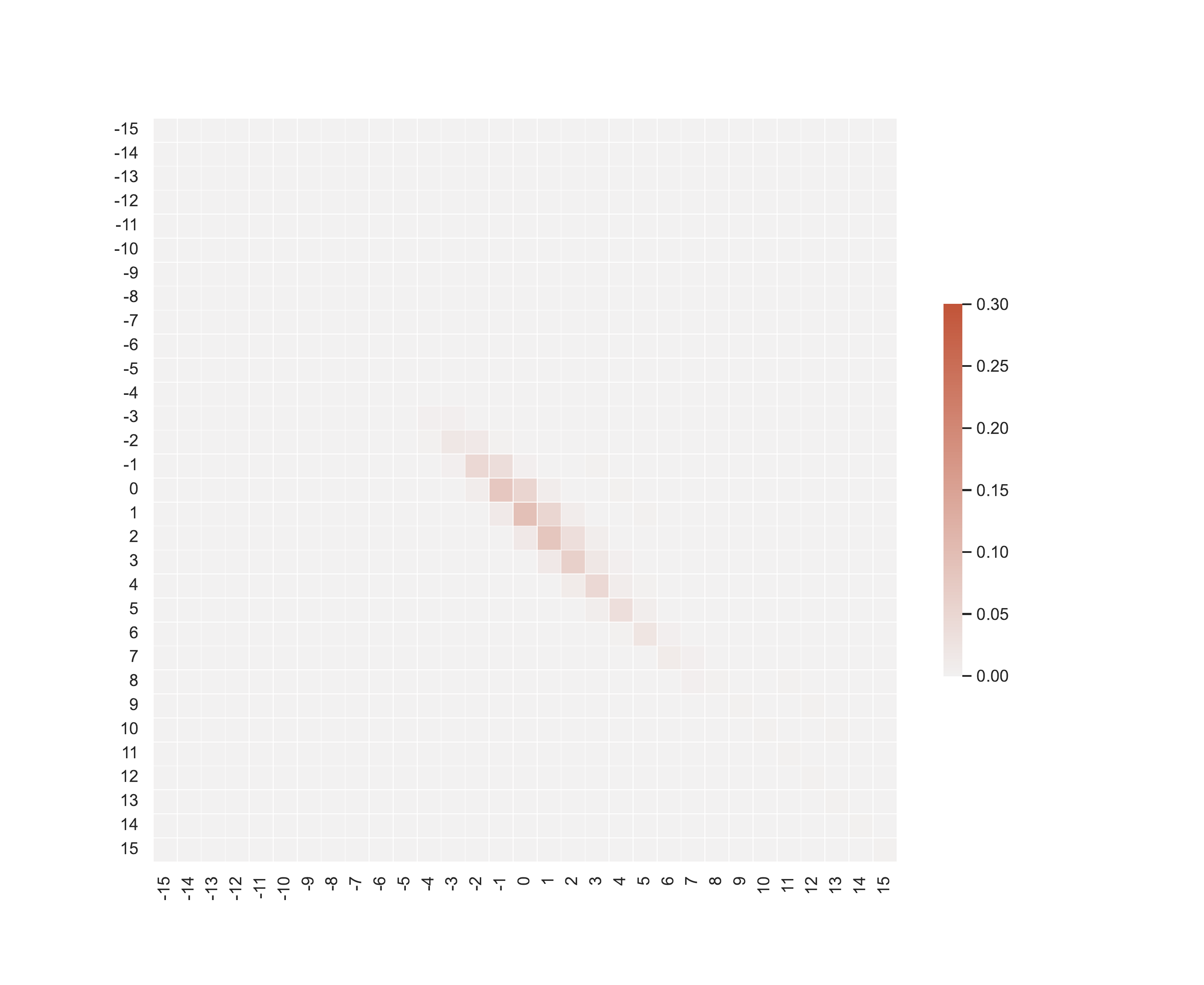}

}
\par\end{centering}
\begin{centering}
\subfloat[Gaussian Kernel Approach]{\includegraphics[scale=0.3]{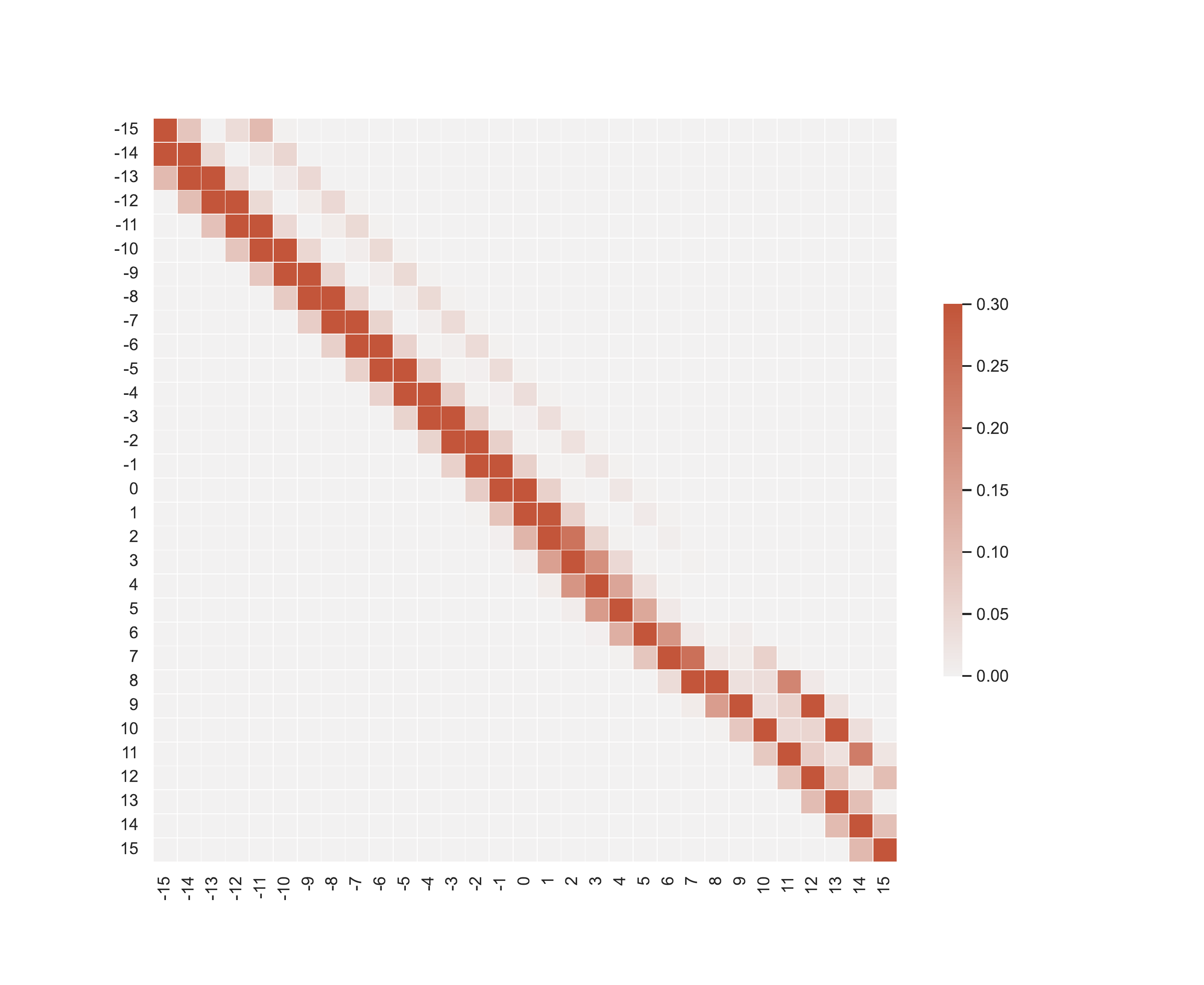}

}
\par\end{centering}
\caption{Gaussian Kernel Method\label{fig:Gaussian-Kernel-Method}}
\end{figure}

\begin{figure}
\begin{centering}
\subfloat[Diagonal Filling Approach]{\includegraphics[scale=0.3]{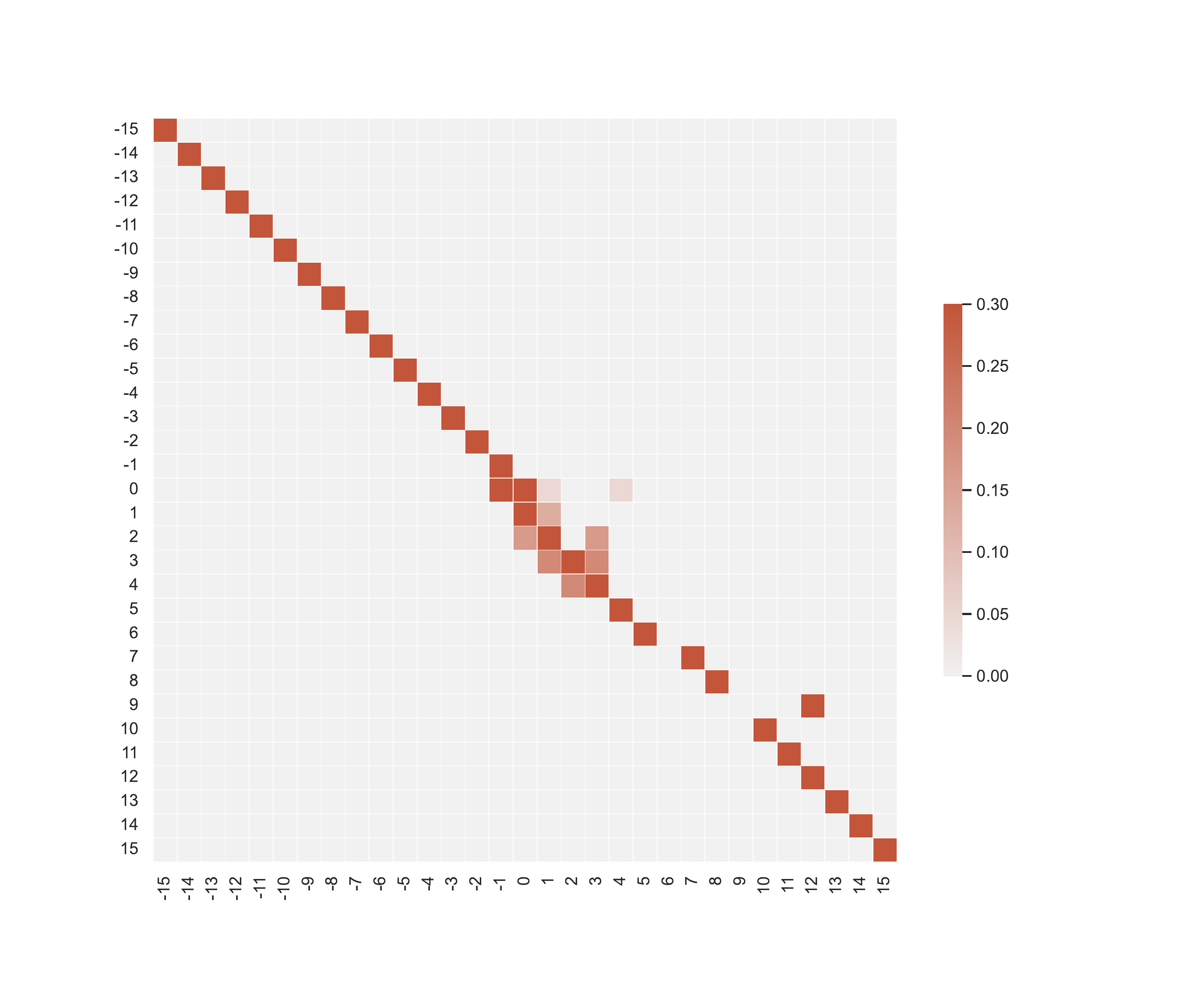}

}\subfloat[Uniform Filling Approach]{\includegraphics[scale=0.3]{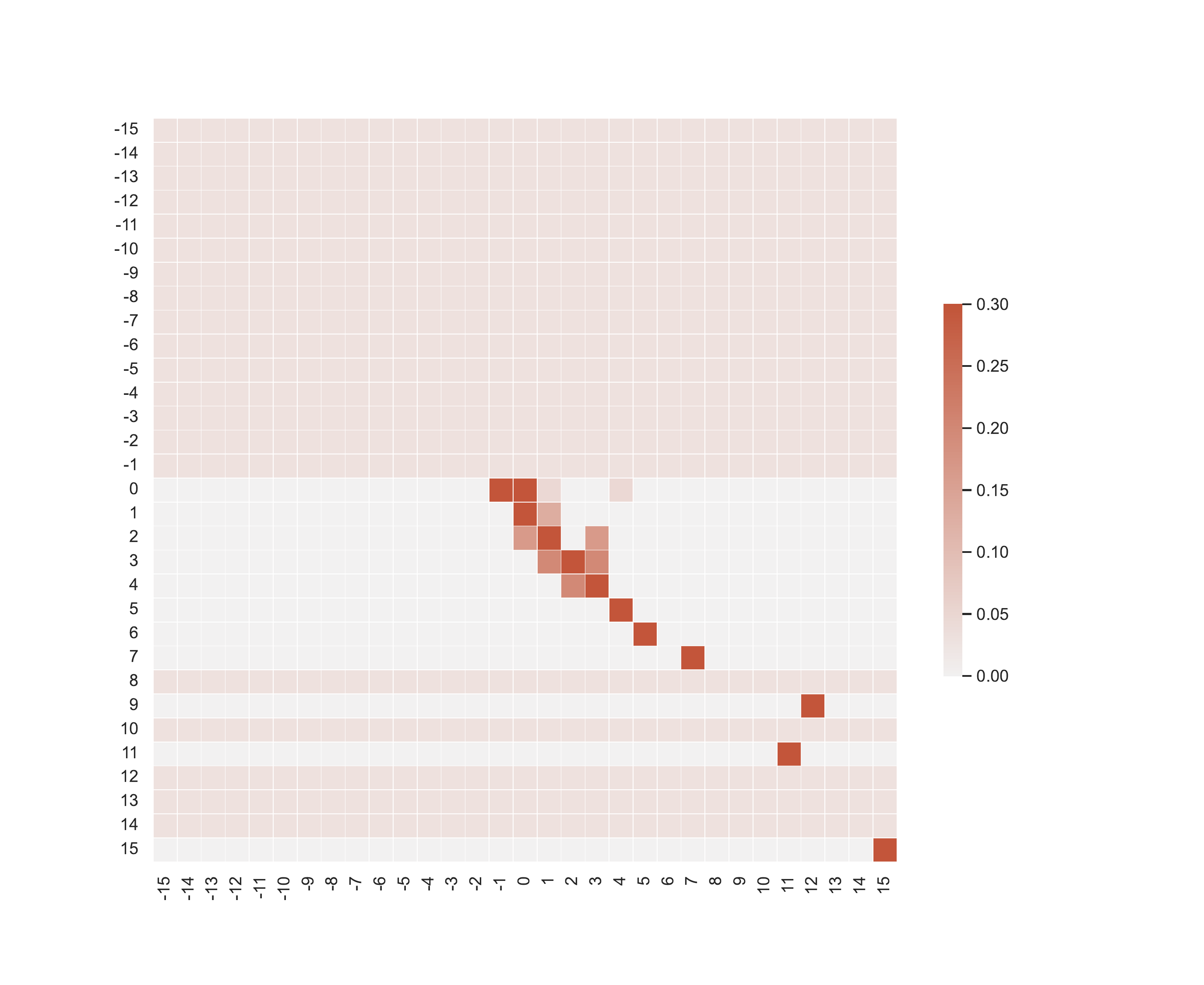}

}
\par\end{centering}
\begin{centering}
\subfloat[Gaussian Regression Approach]{\includegraphics[scale=0.3]{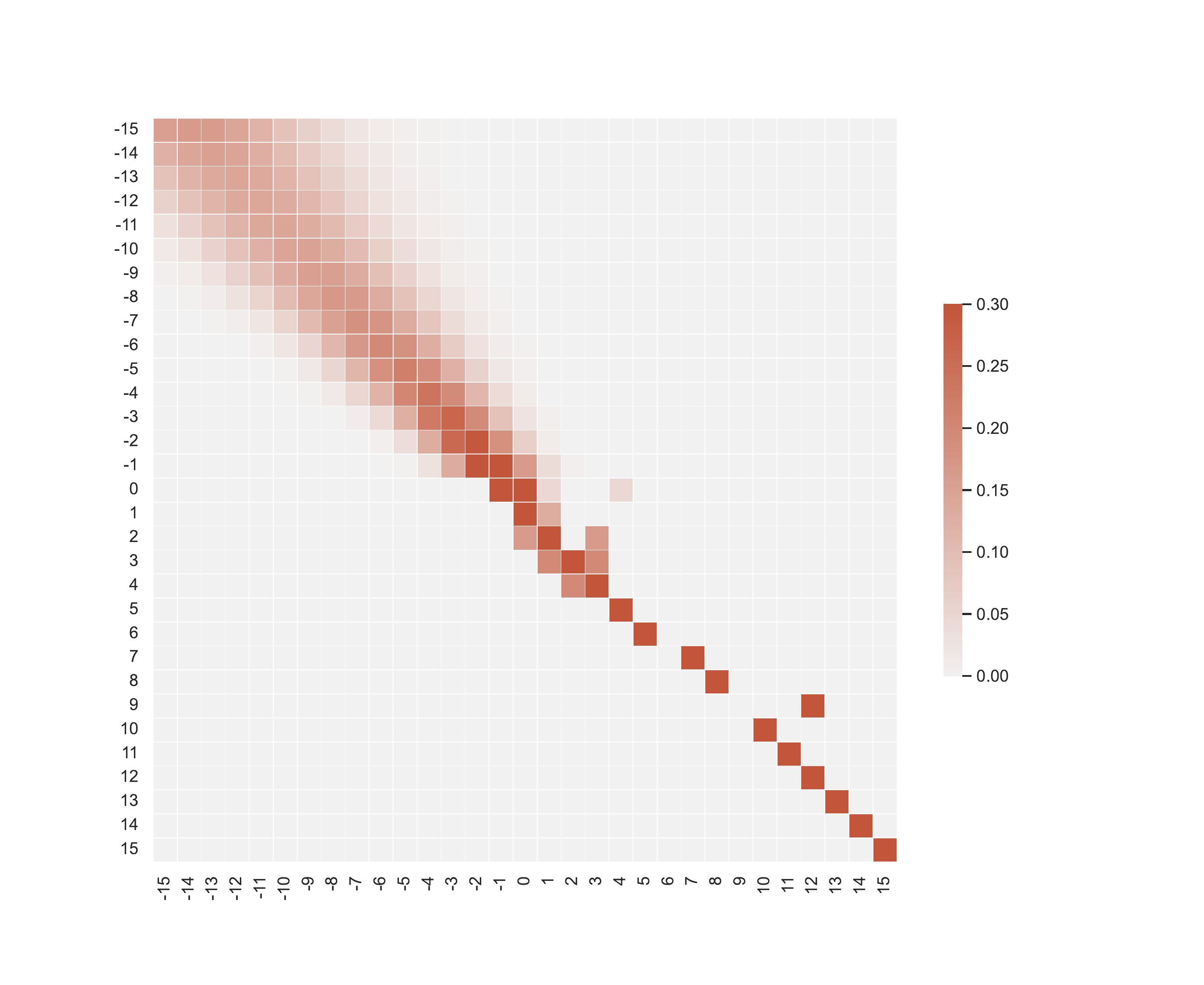}

}
\par\end{centering}
\caption{Comparison of Matrix Recovery Approaches\label{fig:Comparison-of-Matrix}}
\end{figure}

We now compare the matrix recovery approaches discussed above with
the Gaussian kernel approach provided in Section \ref{subsec:The-Gaussian-Kernel-Method}.
We present the Gaussian kernel approach results in Figure \ref{fig:Gaussian-Kernel-Method},
and Figure \ref{fig:Comparison-of-Matrix} further provides the results
for the matrix recovery methods mentioned above. Both Figures \ref{fig:Gaussian-Kernel-Method}
and \ref{fig:Comparison-of-Matrix} are based on the train with the
number ``519'' and station name ``Bl''.

Figure \ref{fig:Gaussian-Kernel-Method}(a) presents the original
transition matrix that we obtained by simply letting $\hat{p}_{j}(t)=\frac{n_{j}(t)}{\sum_{l}n_{l}(t)}$
if $\sum_{l}n_{l}(t)\neq0$. We find that many rows of the transition
matrix are zeros because the negative delay in this station was not
observed. Figure \ref{fig:Gaussian-Kernel-Method}(a) also shows that
the recorded delays are likely to concentrate near the diagonal of
the transition matrix. Only in a few cases do we see that the delay
has a jump. For instance, the delay jumped from 9 minutes to 12 minutes
with a probability 1 in the historical record.

Figure \ref{fig:Gaussian-Kernel-Method}(b) plots the two-dimensional
distribution density after the Gaussian kernel fitting. The delay
density is also concentrated around the diagonal, showing that most
historical delay values are small. Figure \ref{fig:Gaussian-Kernel-Method}(c)
provides the transition matrix recovered by the Gaussian kernel approach.
We obtained Figure \ref{fig:Gaussian-Kernel-Method}(c) by normalizing
each row in Figure \ref{fig:Gaussian-Kernel-Method}(b), so that Figure
\ref{fig:Gaussian-Kernel-Method}(c) is a valid transition matrix
with all values within a row summing to 1. From Figure \ref{fig:Gaussian-Kernel-Method}(c),
we can see that the Gaussian kernel method recovers most features
of the recorded values in Figure \ref{fig:Gaussian-Kernel-Method}(a).
For instance, the recovered transition matrix shows that the next
delay is likely to stay unchanged when the current delay is small. Moreover,
when the current delay is around 9 minutes, the recovered matrix indicates
a high probability that the next delay will jump.

We plot the matrix by diagonal filling approach in Figure \ref{fig:Comparison-of-Matrix}(a),
and the one by uniform filling approach in Figure \ref{fig:Comparison-of-Matrix}(b).
These two approaches do not rely on the available data within the
original transition matrix. We can expect their performance to degrade
when the training data size becomes smaller.

We plot the matrix by the Gaussian regression recovery approach in
Figure \ref{fig:Comparison-of-Matrix}(c). We find that this approach
does not retain the features of the original matrix. Using linear
regression to fit the parameters $\mu_{i}$ and $\sigma_{i}$ can
result in a large $\sigma$ on one side and a small one on the other,
as we observe from Figure \ref{fig:Comparison-of-Matrix}(c). When
the current delay $i$ is small, the fitted $\sigma_{i}$ is large.
Thus we see that the distribution is more dispersed. When the current
delay is large, we even have negative fitted $\sigma_{i}$ values
in the numerical study. For rows with negative fitted $\sigma_{i}$
values, we let its diagonal be 1. So eventually, this approach leads
to a skewed transition matrix, as we see in Figure \ref{fig:Comparison-of-Matrix}(c).

Table \ref{tab:Performance-Comparison-for} provides the performance
comparison for the matrix recovery approaches. Again, we select the
median to predict the delay trend, the jump probability to predict
the delay jump, and the mean to predict the minutes of delay for these
matrix recovery approaches. We find from Table \ref{tab:Performance-Comparison-for}
that the Gaussian kernel method achieves the highest score for delay
jump prediction, RWMSE, and total score, and a similar score in delay
trend prediction as the other filling approaches. This result shows
that the Gaussian kernel recovery approach outperforms the other approaches
in terms of prediction accuracy. Moreover, this result shows that
the good performance of the Gaussian kernel recovery approach is insensitive
to the weights given in Equation (\ref{eq:9}). 

\begin{table}
\begin{centering}
\begin{tabular}{|c|c|c|c|c|c|}
\hline 
Method  & Test Data  & $F_{TR}$  & $F_{JP}$  & RWMSE  & Total Score\tabularnewline
\hline 
\hline 
\multirow{2}{*}{Diagonal Filling} & Test Set 1  & 0.56947  & 0.48485  & 3.04482  & 4.65101\tabularnewline
\cline{2-6} \cline{3-6} \cline{4-6} \cline{5-6} \cline{6-6} 
 & Test Set 2  & 0.71579  & 0.57143  & 2.86607  & 6.42719\tabularnewline
\hline 
\multirow{2}{*}{Uniform Filling} & Test Set 1  & 0.57084  & 0.46877  & 3.04449  & 4.49696\tabularnewline
\cline{2-6} \cline{3-6} \cline{4-6} \cline{5-6} \cline{6-6} 
 & Test Set 2  & \textbf{0.71665}  & 0.48485  & 2.86613  & 5.56580\tabularnewline
\hline 
\multirow{2}{*}{Gaussian Regression} & Test Set 1  & \textbf{0.58557}  & 0.47889  & 5.99850  & 1.72371\tabularnewline
\cline{2-6} \cline{3-6} \cline{4-6} \cline{5-6} \cline{6-6} 
 & Test Set 2  & 0.57463  & 0.38724  & 2.65238  & 4.09323\tabularnewline
\hline 
\multirow{2}{*}{Gaussian Kernel} & Test Set 1  & 0.57231  & \textbf{0.56716}  & \textbf{2.88631}  & \textbf{5.64684}\tabularnewline
\cline{2-6} \cline{3-6} \cline{4-6} \cline{5-6} \cline{6-6} 
 & Test Set 2  & 0.69949  & \textbf{0.62500}  & \textbf{2.58319}  & \textbf{7.16426}\tabularnewline
\hline 
\end{tabular}
\par\end{centering}
\caption{Performance Comparison for Matrix Recovery Approaches\label{tab:Performance-Comparison-for}}
\end{table}

\subsection{Comparison with Other Time Series Models\label{subsec:Comparison-with-Other}}

We now compare our proposed model with other widely-used time series
models.

\subsubsection*{Naive Prediction Approach}

A naive forecasting approach is to assume the delay at station $T$
is identical to the current delay, i.e., $d(T)=d(S)$. This approach
does not rely on historical data, and is simple to implement in reality.

\subsubsection*{Probability Distribution}

We can predict the delay at station $T$ by simply using its delay
distribution. Specifically, we compute $v(T)=(\frac{n_{-N}(T)}{\sum_{l=-N}^{N}n_{l}(T)},...,\frac{n_{N}(T)}{\sum_{l=-N}^{N}n_{l}(T)})$
for station $T$. We then use the median to predict delay trend, use
jump probability to predict delay jump, and delay expectation to predict
the minutes of delay.

\subsubsection*{ARIMA Model}

ARIMA model is a linear time series forecasting model that has been
used in various forecasting fields, including economics, engineering,
and geology \citep{cryer2008time}. Originated from the autoregressive
(AR) model and the moving average (MA) model, the ARIMA model can
be used when the time series is stationary and with no missing values
\citep{ediger2007arima}. An ARIMA model is usually written as $ARIMA(p,\beta,q)$,
where $p$ is the order of the AR term that refers to the number of
lags to be used as predictors. The parameter $q$ is the order of
the MA term that refers to the number of lagged forecast errors that
should go into the ARIMA model. The parameter $\beta$ is the minimum
number of differencing needed to make the series stationary.

We implement a nonseasonal ARIMA model with the Python package ``pmdarima''
\citep{pmdarima} to predict the delay. We now use \emph{Test Set 1}
as an example to demonstrate how we develop the ARIMA model. For each
train, we use the delays recorded before 8:00 as the training set
for the ARIMA model. Then we use the trained model to predict the
delay at the predicted station. A detailed description of how to use
ARIMA model to predict several time units into the future can be found
in \citet{cryer2008time}.

We conduct a stepwise algorithm \citep{hyndman2008automatic} to find
the optimal model parameters, including the order of AR $p\in\{1,2,3\}$
and the order of MA $q\in\{1,2,3\}$. We rely on the Akaike's Information
Criterion (AIC) value in selecting the best orders of the ARIMA model.
We further obtain the optimal value $\beta$ by conducting differencing
tests.

\subsubsection*{Model Comparison}

We compare the prediction scores of the time series model mentioned
above in Table \ref{tab:time series-Model}. Although the naive forecasting
approach has the minimum RWMSE, its $F_{TR}$ and $F_{JP}$ scores
are the worst since it does not utilize the historical data for training.
Using probability distribution to predict delay can be promising when
the delay is stable. For instance, its performance for \emph{Test
Set 2} is close to the performance of our Markov chain model. However,
for \emph{Test Set 2} whose delays are divergent and unstable, the
performance of using probability distribution is much worse than our
model. The reason is that the probability distribution is only obtained
from historical data. This approach does not utilize the delays at
the past stations on the particular predicted date.

The ARIMA model only uses the historical data on the predicted date
for training without using the other historical data. When the predicted
train's current station $S$ is not far from its starting station
1, the training set could be small so that not much pattern can be
learned from history. We can find from Table \ref{tab:time series-Model}
that the ARIMA model performs worse than the method of using probability
distribution due to its training set being small.

The Markov chain model we proposed in this work has a better overall
performance than the other models described above. The reason is that
the Markov chain model is trained based on historical data and also
takes the delay at the current station as the input. Our model has
the largest score in $F_{TR}$. Its $F_{JP}$ score is the best for
\emph{Test Set 1} where the delays are divergent. Its RWMSE scores
are close to the minimal scores given by the naive forecasting approach.
We can then conclude that the good performance of our model is not
sensitive to the weights assigned in Equation (\ref{eq:9}). 

\begin{table}
\begin{centering}
\begin{tabular}{|c|c|c|c|c|c|}
\hline 
Method  & Test Data  & $F_{TR}$  & $F_{JP}$  & RWMSE  & Total Score\tabularnewline
\hline 
\hline 
\multirow{2}{*}{Naive Forecasting} & Test Set 1  & 0.18156  & 0.00476  & \textbf{2.86511}  & -1.95253\tabularnewline
\cline{2-6} \cline{3-6} \cline{4-6} \cline{5-6} \cline{6-6} 
 & Test Set 2  & 0.24182  & 0.00111  & \textbf{2.45632}  & -1.23613\tabularnewline
\hline 
\multirow{2}{*}{Probability Distribution} & Test Set 1  & 0.43879  & 0.53488  & 4.00108  & 3.54168\tabularnewline
\cline{2-6} \cline{3-6} \cline{4-6} \cline{5-6} \cline{6-6} 
 & Test Set 2  & 0.64149 & \textbf{0.64849} & 2.58138 & 7.11094\tabularnewline
\hline 
\multirow{2}{*}{ARIMA Model} & Test Set 1  & 0.39252  & 0.28000  & 3.44114  & 1.32148\tabularnewline
\cline{2-6} \cline{3-6} \cline{4-6} \cline{5-6} \cline{6-6} 
 & Test Set 2  & 0.67438  & 0.42424  & 2.60824  & 5.00610\tabularnewline
\hline 
\multirow{2}{*}{Our Model} & Test Set 1  & \textbf{0.57231}  & \textbf{0.56716}  & 2.88631  & \textbf{5.64684}\tabularnewline
\cline{2-6} \cline{3-6} \cline{4-6} \cline{5-6} \cline{6-6} 
 & Test Set 2  & \textbf{0.69949}  & 0.62500  & 2.58319  & \textbf{7.16426}\tabularnewline
\hline 
\end{tabular}
\par\end{centering}
\caption{Time Series Model Comparison \label{tab:time series-Model}}
\end{table}

\section{Conclusion and Future Research\label{sec:Conclusion-and-Future}}

This research proposes a novel, accurate, and efficient model to predict
railway delays using a Markov-chain-based framework. Theoretical properties
of the proposed model are rigorously investigated, and insights are
developed. Moreover, we conduct numerical experiments to verify the
prediction accuracy and efficiency using the Netherlands Railways
data. The major findings of this paper are summarized below: 
\begin{itemize}
\item We propose a non-homogeneous Markov chain to characterize the delay
process over stations. To test the order of the Markov chain, we propose
and conduct a chi-square Markov property test with the Netherlands
Railways data. The results show that the delays over stations of the
same train follow a first-order Markov chain. 
\item We develop a Gaussian-kernel-based method to recover the transition
matrices for the Markov chain model when the size of training data
is small. The Markov chain model equipped with the proposed recovery
method achieves a higher prediction accuracy than being with other
heuristic matrix recovery methods. 
\item We conduct numerical experiments on the real-world Netherlands Railways
data and measure the prediction performance of each implemented model
using a certain score calculated with the delay trend, the delay jump,
and the root weighted mean square error. The proposed model provides
a higher prediction score than other benchmark time series prediction
models. 
\end{itemize}
Our investigation of the Markov-chain-based delay prediction model
leads to some potential research works. For instance, the strategies
for train scheduling and passenger assignment can be improved when
considering the potential railway delay \citep{xu2019simultaneous,li2020optimizing}.
Incorporating delay prediction into online railway scheduling and
dispatching is one of our future research directions. Besides,
we can migrate the proposed Markov property test, Markov chain model, and matrix
recovery method to other application areas, such as
health prognostics \citep{ghamlouch2018prognostics} and electric load
prediction \citep{khashei2021comprehensive}.

\printbibliography

\end{document}